# SAGE-XGBoost: Spatially Augmented Graph Embeddings–Machine Learning Framework for Natural Hazards Susceptibility Mapping under Data Scarcity

Mohammad H. Vahidnia[1,*], Ali Pourkarimi[1]

[1] *Center for Remote Sensing and GIS Research, Faculty of Earth Sciences, Shahid Beheshti University, Tehran, Iran*

mh_vahidnia@sbu.ac.ir; a.pourkarimi@mail.sbu.ac.ir

**Abstract**

Natural hazard susceptibility mapping is often constrained by limited labeled data, reducing the generalizability of conventional machine learning and limiting the applicability of complex deep learning models. This study proposes SAGE (Spatially Augmented Graph Embeddings), a structurally informed feature-engineering framework that combines controlled noise-based data augmentation with neighborhood-based graph embeddings to improve prediction under data-scarce conditions. A K-nearest neighbor graph is constructed to derive local spatial statistics, which are reduced using principal component analysis and integrated with environmental covariates and spatial coordinates. The resulting features are used with XGBoost to develop the SAGE-XGBoost model. The framework was evaluated for landslide and wildfire susceptibility mapping. SAGE-XGBoost consistently outperformed conventional and spatially explicit machine learning models. Compared with Spatial XGBoost, it achieved an absolute improvement of above 33 percentage points across the two case studies. The model reached AUC values of approximately 0.97 for landslide susceptibility and 0.95 for wildfire susceptibility. Feature importance analysis confirmed the contribution of graph embeddings to prediction, while their integration improved spatial coherence and reduced local noise amplification. Overall, SAGE-XGBoost provides an efficient and transferable alternative to deep representation learning for environmental hazard assessment and other geospatial prediction tasks under limited supervision.

**Keywords:** Structurally Informed Learning, GeoAI, Spatial Heterogeneity, Landslide Susceptibility Mapping, Wildfire Susceptibility Mapping, Spatial Machine Learning, Graph Embeddings, Data Augmentation

## 1. Introduction

In recent decades, the increasing losses attributed to natural hazards such as landslides, floods, earthquakes, and soil erosion have driven researchers and policymakers toward advanced tools for identifying and assessing vulnerable areas. In this context, Geographic Information Systems (GIS), as powerful platforms for the collection, management, analysis, and visualization of spatial data, are pivotal to the development of susceptibility maps (Vahidnia et al., 2010; Tehrany et al., 2015). By utilizing GIS, multiple layers of spatial information can be integrated, enabling multi-dimensional analyses of floods, earthquakes, landslides, storms, and other natural hazards. Accurate spatial mapping and hazard modeling facilitate the production of vulnerability and risk maps, which provide an essential basis for optimal decision-making in disaster management, risk reduction, and resource planning (Dano et al., 2019; Tacconi Stefanelli et al., 2020).

The process of susceptibility mapping—also referred to as susceptibility modeling or susceptibility assessment—

[*] *Corresponding Author*

involves the collection and analysis of a set of factors influencing the occurrence of a given phenomenon (Abedi Gheshlaghi et al., 2021; Thiery et al., 2007; Ribeiro et al., 2017). For instance, in landslide studies, variables including slope gradient, lithology, land use, precipitation, distance to faults and rivers, and vegetation cover are of major importance. Data related to these factors are commonly obtained from topographic maps, satellite imagery, field surveys, and meteorological records. Subsequently, each factor is weighted according to its relative influence on the geological phenomenon, and statistical, machine learning, or other modeling approaches are applied to generate the final susceptibility map. Beyond applications in landslides, wildfires, and floods, such approaches have been widely adopted in studies of avalanches, debris flows, rockfalls, droughts, and many other destructive phenomena (Mosavi et al., 2020; Kumar et al., 2016; Gao et al., 2025; Li et al., 2024; El Miloudi et al., 2025). Ultimately, susceptibility maps enable the identification of regions with the highest potential for the occurrence of hazardous events. Susceptibility mapping generally refers to the delineation of areas prone to a specific phenomenon, without explicitly considering the temporal dimension. This concept differs from *hazard*, which represents the probability of occurrence at a particular location and time, and from *vulnerability*, which reflects the susceptibility of society and infrastructure to damage (Van Westen et al., 2008; Ha et al., 2023).

Approaches such as statistical approaches, machine learning, multi-criteria decision analysis (MCDA), hybrid methods, and lately deep learning are among the most widely used techniques in this field (Azarafza et al., 2021; Ado et al., 2022; Nachappa et al., 2020; Jahanbani et al., 2024). The selection of an appropriate method depends on data characteristics, study objectives, and regional conditions, and is typically determined through an evaluation process using inventory data. Finally, the production of classified susceptibility maps (e.g., low, moderate, high, and very high) facilitates more straightforward decision-making regarding areas most prone to hazards.

Statistical methods represent some of the earliest techniques employed in susceptibility mapping. These approaches are based on analyzing the relationships between event occurrences (e.g., landslides or floods) and environmental conditioning factors. Techniques such as Weight of Evidence and Frequency Ratio estimate occurrence probabilities in a binary manner, whereas models such as Logistic Regression allow for multivariate analysis and quantification of the contribution of each factor (Lee, 2004; Ilia and Tsangaratos, 2016; Rahmati et al., 2016). The primary advantages of these techniques are their simplicity, transparency, and lower data requirements; however, their limitations include reduced capability to model complex nonlinear relationships and limited generalization performance (Reichenbach et al., 2018). Other statistical approaches, such as Certainty Factor (CF) and Information Value (IV), have also been applied in previous studies (Ba et al., 2017; Chen et al., 2016).

With the expansion of spatial datasets and advances in data-mining algorithms, machine learning techniques have been extensively utilized for susceptibility assessment. Techniques such as Random Forest, Decision Tree, Support Vector Machine, Gradient Boosting Tree, and Artificial Neural Networks (ANN) are capable of identifying nonlinear relationships among conditioning factors and providing high predictive accuracy while adapting to complex datasets (Zhao et al., 2021; Gigović et al., 2019). Compared to traditional statistical models, these approaches are more flexible and better suited for processing large and heterogeneous datasets (Zhao et al., 2022).

More recent studies have demonstrated that ensemble algorithms based on boosting and bagging techniques—particularly Gradient Boosting Trees and XGBoost—are among the most effective methods in this domain (Can et al., 2021; Abedi et al., 2022). Although advanced models play a crucial role, their performance does not rely solely on algorithmic complexity. In this regard, a comprehensive review by Pourghasemi et al. (2018) highlighted that the selection of input variables, including topographic, geological, land cover, and hydrological parameters, is a determining factor in the quality of susceptibility maps. Furthermore, recent studies emphasize the integration of data from remote sensing with GIS to improve model accuracy (Pugliese Viloria et al., 2024).

A growing body of research has shifted toward hybrid models that combine data-driven algorithms with knowledge-based approaches. These models leverage the strengths of both categories and have demonstrated improved accuracy in many studies (Pugliese Viloria et al., 2024). For example, optimization techniques, including Genetic Algorithms, Ant Colony, and Particle Swarm Optimization (PSO), are utilized to enhance and fine-tune the parameters of machine learning models. Additionally, combining MCDM with machine learning aids in optimal variable selection and enhances predictive performance. Neuro-fuzzy approaches, in particular, have shown strong capabilities in integrating expert knowledge with data in susceptibility mapping. Recent reviews have also identified spatial validation and model generalizability to new geographic domains as major research gaps in this field (Reichenbach et al., 2018).

Deep learning approaches particularly Convolutional Neural Networks (CNNs) and their derivatives have been applied to image-based analyses within susceptibility mapping studies in recent years (Ngo et al., 2021; Youssef et al., 2022). These image-based data are primarily used during data preparation stages and for detecting past events as evidence for supervised learning. Moreover, although susceptibility mapping is conventionally defined as identifying prone areas by aggregating information across different time periods without explicitly considering time as an independent dimension, recurrent neural networks (RNNs), Long Short-Term Memory (LSTM), crossbreed CNN–LSTM architectures, and attention-based models—particularly transformers—have recently been introduced into susceptibility modeling for time-series data (Yan et al., 2025; Anil and Manjula, 2025). These methods can automatically extract spatiotemporal features and are particularly powerful for processing satellite imagery and multi-source datasets. Nevertheless, their reliance on extensive datasets and substantial computational resources remains a key limitation.

Other emerging approaches include methods that integrate spatial statistics with machine learning models, such as Spatial XGBoost and Geographically Weighted Random Forest, which have often demonstrated improved modeling performance. In addition, explainable AI techniques, such as SHAP, have proven highly efficacious for interpreting variable importance and influence (Lu et al., 2025; Pradhan et al., 2023).

One of the most critical challenges in GIS-based susceptibility mapping, however, is data quality. Incomplete or heterogeneous datasets can substantially reduce model accuracy, generalizability, and predictive capability. In many susceptibility mapping studies across landslides, floods, wildfires, avalanches, and similar hazards, researchers encounter regions with limited labeled samples of past events (Falah et al., 2019; Kong et al., 2025; Chihi et al., 2025; Kanani-Sadat et al., 2019). In other words, data scarcity arises from limited historical records, sparse inventories, and incomplete environmental observations.

Data scarcity restricts the applicability of highly complex models such as deep learning and often leads to suboptimal performance even for advanced machine learning and hybrid approaches. Previous studies have largely focused on knowledge-based or unsupervised methods rather than leveraging the available data to enhance model performance. Moreover, most proposed solutions are domain-specific, and only a limited number of studies have addressed the development of a comprehensive framework applicable across various natural hazards susceptibility mapping problems. This research seeks to introduce a novel, fully data-driven, and general framework aimed at improving natural hazards susceptibility mapping in data-scarce regions.

It is noteworthy to mention that the concept of data scarcity—referred to in related contexts as limited inventory, scarce historical records, or sparse event data—lacks a standardized definition, and no specific threshold or spatial unit can be universally defined for minimum sample requirements.

In this study, we define data-scarce conditions based on the limited density and spatial coverage of hazard inventories relative to the extent of the study area and the complexity of the predictor space. Specifically, the landslide inventory in the one case study contains only 191 recorded events, while the wildfire inventory in the second case study contains 128 events, despite both study areas comprising millions of 30-m grid cells. It is clear that these inventories provide a limited representation of the spatial and environmental characteristics that exist throughout the landscape. As a result, many predictor variable combinations are either completely absent or underrepresented in the training set. Therefore, the term "data scarcity" in this study refers to insufficient spatial and feature-space coverage of hazard occurrences. Consequently, due to the infeasibility of employing models such as CNNs, LSTMs, and transformers, and the inadequate performance of conventional machine learning models in attaining effective generalization, we pursue a third alternative: a model capable of achieving high-quality results from limited hazard inventories.

The proposed approach is termed SAGE, with specially focusing on SAGE-XGBoost, where SAGE stands for *Spatially Augmented Graph Embeddings* and XGBoost is the classifier that implements it in this study. The proposed framework is built upon three key components: data augmentation, extraction of spatial features in the form of graph-based neighborhood information, and embedding generation through dimensionality reduction to capture the most informative spatial feature components. While most existing studies rely solely on direct point-based features such as elevation, NDVI, slope, wind, rainfall, and vegetation type, we argue that modeling spatial heterogeneity requires incorporating graph-neighborhood features, including mean, weighted mean, standard deviation, density, and related statistics. Furthermore, standard data augmentation techniques are employed to generate synthetic samples similar to real observations, thereby enhancing model learning.

In recent years, spatial and graph embeddings have emerged as powerful approaches for extracting information from point neighborhoods and structured networks (Belussi et al., 2024; Du et al., 2022); however, their direct application in susceptibility mapping using machine learning has received limited attention. It is important to note that the GraphSAGE (SAmple and aggreGatE) architecture (Hamilton et al., 2017) is not related to the proposed SAGE-XGBoost framework. SAGE-XGBoost employs a graph-based feature engineering technique to obtain neighborhood statistics and low-dimensional graph embeddings prior to XGBoost classification, whereas GraphSAGE is an inductive graph neural network intended for representation learning through neighborhood aggregation. Therefore, neither message-passing operations nor graph neural network training are used in the suggested architecture.

By introducing the SAGE-XGBoost framework, this study seeks to assess the effectiveness of these techniques under data-scarce conditions. To this end, two case studies—one related to landslides and the other to wildfires—are conducted to validate the performance and robustness of the proposed method.

## 2. Methodology

In this study, a novel graph-embedded spatial machine learning (SAGE-XGBoost) framework is proposed for natural hazards susceptibility mapping under conditions of limited labeled data. The framework integrates environmental covariates, data augmentation, spatial relationships, and neighborhood-based structural information to enhance model generalization, robustness, and spatial consistency. **Figure 1** represents the workflow of SAGE-XGBoost framework. The subsequent subsections will delineate the intricacies of each process.

The novelty of the proposed framework does not stem from the development of entirely new algorithms. Rather, it arises from the systematic integration of established techniques—including KNN-based neighborhood analysis, PCA, data augmentation, and XGBoost—into a unified spatial learning framework tailored for susceptibility mapping under data-scarce conditions. The contribution therefore lies in the design and effective combination of these components to address a specific and challenging application setting.

The proposed SAGE-XGBoost framework integrates graph-derived neighborhood descriptors, low-dimensional spatial embeddings, and controlled sample augmentation to address spatial dependency and limited inventory size concurrently. In contrast to graph neural network approaches, the framework captures local spatial structure without requiring graph training or message-passing operations, thus ensuring both computational efficiency and ease of application to conventional tabular susceptibility datasets.

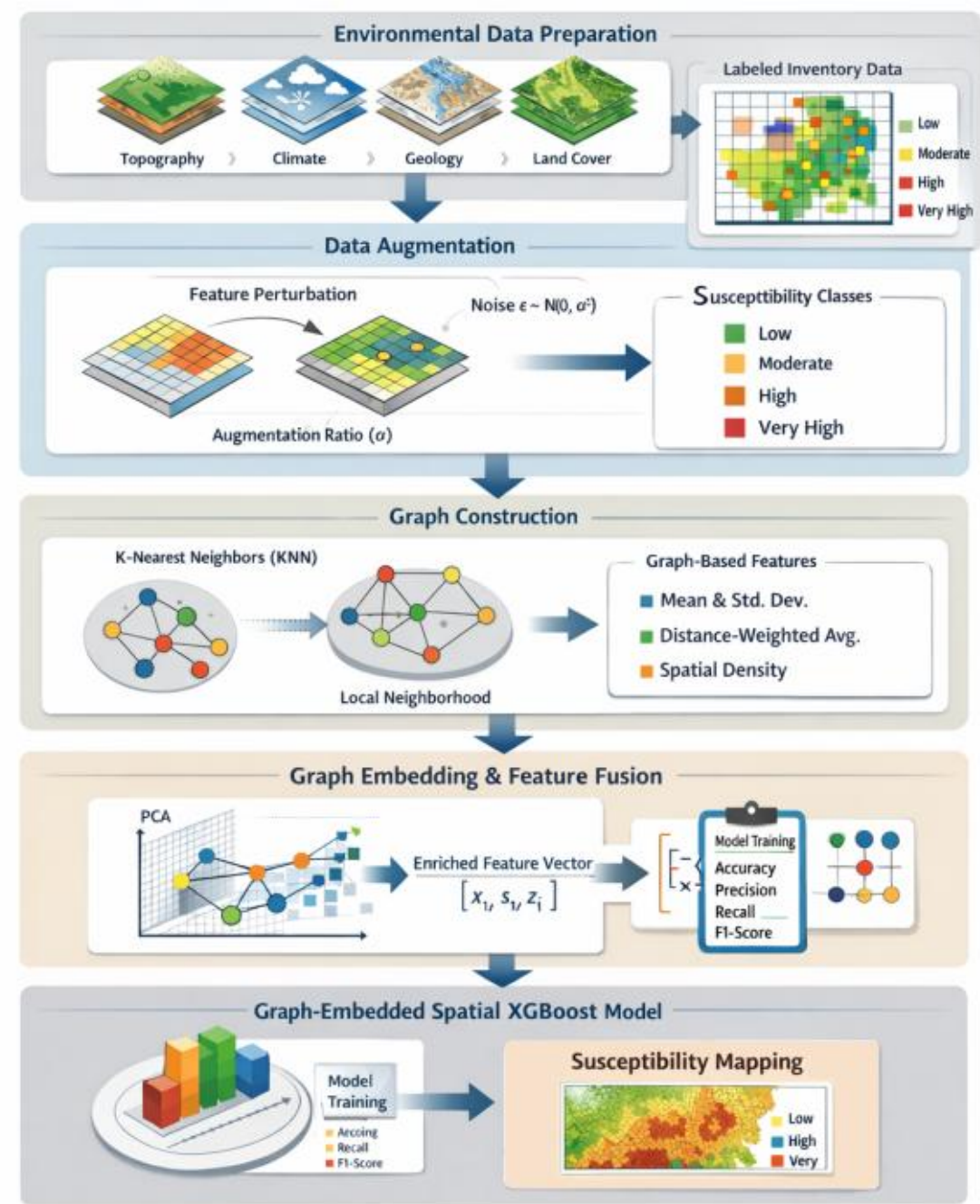

Figure 1. The methodology of SAGE-XGBoost (Spatially Augmented Graph Embeddings XGBoost) framework for natural hazards susceptibility mapping

### 2.1. Inventory Data and Labeling

Initially, depending on the target susceptibility problem, multiple environmental conditioning factors—including topographic, climatic, geological, and land-cover variables—are prepared as raster layers and spatially co-registered. Let $\mathbf{X} = \{X_1, X_2, \dots, X_m\}$ denote the set of $m$ environmental raster layers (e.g., topographic, climatic, geological, and land-cover variables). Each spatial location $i$ is represented by a feature vector:

$$\mathbf{x}_i \in \mathbb{R}^m \quad (1)$$

Labeled inventory data (e.g., landslide or wildfire occurrences) are then incorporated to construct the response variable. In this study, inventory samples are categorized into three susceptibility classes based on the observed intensity of impact or destruction. These classes are further complemented by an additional background class representing non-inventory locations, corresponding to areas with low or no susceptibility. Consequently, the final classification task is formulated as a four-class problem:

$$y_i \in \{0,1,2,3\} \quad (2)$$

### 2.2. Data Augmentation under Limited Labeled Samples

To mitigate the effects of data scarcity, an optional data augmentation strategy based on controlled feature perturbation ($\mathcal{E}$) is applied. For each original sample $\mathbf{x}_i$, augmented samples are generated as:

$$\tilde{\mathbf{x}}_i = \mathbf{x}_i + \mathcal{E}, \quad where\ \mathcal{E} \sim \mathcal{N}(0, \sigma^2) \tag{3}$$

Here, the noise standard deviation is defined separately for each feature (j) as:

$$\sigma_j = \alpha SD_j \tag{4}$$

Where $SD_j$ denotes the empirical standard deviation of feature $j$, and $\alpha$ is a scaling factor controlling the perturbation magnitude to preserve the physical meaning of environmental variables. In this study, $\alpha$ = 0.05, yielding $\sigma_j = 0.05\, SD_j$. The total number of augmented samples is defined as:

$$N_{aug} = \alpha . N \tag{5}$$

where $N$ denotes the number of original samples and $\alpha \in (0,2)$ signifies the augmentation ratio. The optimal augmentation ratio is determined empirically by incrementally increasing $\alpha$ until classification accuracy improves by less than 1%, thereby preventing overfitting and redundancy.

It should be noted that applying augmentation exclusively to the training data represents the preferred practice than applying it on the total dataset. To avoid data leakage in the SAGE-XGBoost framework, after non-hazard events (non-landslides, or non-wildfire events) are added to the inventory samples, the data is split into test and training. After that, the augmentation is added at a controlled rate to the training data, not the test data.

## 2.3. Graph Construction and Neighborhood Feature Extraction

To explicitly model spatial context, a spatial graph $G = (V, E)$ is built using a K-nearest neighbor (KNN) strategy, where each node corresponds to a labeled sample and edges connect spatially proximate samples. Unlike fixed-radius graphs, the KNN approach ensures adaptive neighborhood definition across heterogeneous spatial densities. For each node $i$, its neighborhood $\mathcal{N}_i$ consists of the $K$ nearest samples in Euclidean space. During implementation, we can normalize the distances to the [0, 1] range using the dimensions of the corresponding study area, thereby defining a scale-independent neighborhood relationships. Based on this neighborhood, a set of graph-based features is extracted, including:

- Neighborhood mean:

$$\boldsymbol{\mu}_i = \frac{1}{K} \sum_{j \in \mathcal{N}_i} \mathbf{x}_j \tag{6}$$

- Neighborhood standard deviation:

$$\boldsymbol{\sigma}_i = \sqrt{\frac{1}{K} \sum_{j \in \mathcal{N}_i} \left(\mathbf{x}_j - \boldsymbol{\mu}_j\right)^2} \tag{7}$$

- Distance-weighted mean:

$$\boldsymbol{\mu}_i^{(w)} = \frac{\sum_{j \in \mathcal{N}_i} w_{ij} \mathbf{x}_j}{\sum_{j \in \mathcal{N}_i} w_{ij}}, \ w_{ij} = \frac{1}{d_{ij} + \varepsilon} \tag{8}$$

Here, $d_{ij}$ denotes the Euclidean distance between nodes $i$ and $j$.

- Local spatial density:

$$\rho_{ij} = \frac{1}{\frac{1}{K} \sum_{j \in \mathcal{N}_i} d_{ij} \quad + \varepsilon} \tag{9}$$

These features collectively characterize neighborhood heterogeneity and local spatial structure. Although alternative graph construction strategies can be employed, the KNN-based approach is sufficient for susceptibility mapping problems, where node degree is assumed to be approximately constant and spatial proximity dominates interaction strength. In this study, the maximum number of neighbors $K$ is varied between 5 and 20 to assess sensitivity.

Spatial neighborhoods were constructed using Euclidean distances computed in a normalized raster-coordinate space. The dimensions of the research area were used to scale each sample's raster row and column indices to the interval [0,1]. Consequently, neighborhood associations were established irrespective of each case study's absolute spatial extent, thereby enabling the construction of comparable graphs across datasets of varying sizes and geometries. A tree-based implementation was utilized to ascertain a fixed number K of nearest neighbors for each node. No explicit boundary correction was applied; samples located near study-area boundaries were connected to the nearest available neighbors within the dataset. Due to the fact that K was fixed for all samples, isolated or degenerate neighborhood configurations were avoided.

## 2.4. Graph Embedding and Feature Fusion

To obtain a compact representation of the neighborhood structure, principal component analysis (PCA) is employed to generate low-dimensional graph embeddings. These embeddings are fused with the original environmental features and normalized spatial coordinates, forming an enriched feature space. The PCA is applied as:

$$\boldsymbol{z}_i = \mathbf{P}^{\mathrm{T}} . \mathbf{g}_{\mathrm{i}} \tag{10}$$

where $\mathbf{g}_i$ is the graph feature vector and $\boldsymbol{z}_i$ is the low-dimensional embedding. The number of retained components is varied between 2 and 5, and the optimal configuration is selected based on predictive performance. The final feature vector for each sample is constructed by concatenating:

$$\mathbf{f}_i = [\mathbf{x}_i, \mathbf{s}_i, \mathbf{z}_i] \quad (11)$$

where $\mathbf{s}_i$ denotes normalized spatial coordinates.

### 2.5. Model Training, Evaluation, and Mapping

The enriched feature set is used to train a SAGE-XGBoost model, where XGBoost is preferred for its robustness, scalability, and proven performance in environmental susceptibility studies. Model performance is evaluated using precision, accuracy, F1-score, and recall, and compared against baseline ML models (Support Vector Machine, Random Forest, and conventional XGBoost), as well as spatially explicit approaches including Geographically Weighted Random Forest (GWRF) and Spatial XGBoost. Finally, the trained model is applied to all valid raster pixels within the study area to generate continuous susceptibility maps. The proposed framework is validated through two case studies—landslide susceptibility mapping and forest fire susceptibility mapping—demonstrating its effectiveness, adaptability, and general applicability across different environmental hazards.

Graph-based features are not restricted to the inventory samples utilized in model training for susceptibility mapping. Rather, each valid raster pixel in the study region has its graph descriptors recalculated. In particular, the same neighborhood statistics (neighbor mean, neighbor standard deviation, distance-weighted mean, and local density) used during training are computed for each pixel after the K nearest inventory samples were found in the normalized spatial coordinate space. To get graph embeddings for every raster pixel, the resulting graph feature vectors are then projected into the PCA space. Consequently, graph embeddings are explicitly generated for the entire study area rather than interpolated from training samples, ensuring consistency between the training and mapping stages.

## 3. Data Preparation

### 3.1. Case Study of Landslide Susceptibility Mapping (LSM)

The first case study focuses on landslide susceptibility in Ardabil County. Ardabil County is located in northwestern Iran, within Ardabil Province, and lies approximately between 37.15° to 38.62° N latitude and 47.7° to 48.9° E longitude (**Figure 2**). The county is situated on the slopes of Mount Sabalan and is characterized by mountainous terrain, steep hillslopes, and intermontane plains. The region's lithological diversity, comprising Eocene to Quaternary volcanic units, recent alluvial deposits, and varied sedimentary formations, in conjunction with relatively sufficient precipitation (characterized by a cold semi-humid climate), an active drainage network, and diverse land-use patterns, engenders conditions conducive to slope instability and landslide occurrence.

In this study, eleven criterion maps were prepared as input variables, including aspect, climate (a composite index of precipitation and humidity), distance to river, elevation, distance to road, land use, distance to fault, lithology, slope, soil, and vegetation. To generate the landslide-conditioning factor layers, a combination of remote sensing data and ground-based sources was utilized. The Digital Elevation Model (DEM) was derived from SRTM products with a 30-meter spatial resolution, and the slope, aspect, and elevation layers were extracted using analytical functions within the Geographic Information System (GIS) environment. Fault networks and lithological units were obtained from 1:100,000 maps of the Geological Survey Organization of Iran (https://gsi.ir/en). These maps were converted into raster format. The distance-to-fault, distance-to-river, and distance-to-road layers were produced using the technique of Euclidean Distance analysis, which was based on corresponding vector maps (available datasets).

The derivation of land-use and vegetation layers was accomplished through the utilization of Landsat 8 OLI satellite imagery (https://earthexplorer.usgs.gov/), which boasts a spatial resolution of 30 meters. This product was supplied by the National Cartographic Center (https://en.ncc.gov.ir/). The soil layer was prepared from digital soil maps at a scale of 1:100,000 and subsequently reclassified and converted into a 30-meter raster format. The climatic index was generated using precipitation and relative humidity data from regional meteorological stations and interpolated spatially using the Inverse Distance Weighting (IDW) method. It was then normalized as a composite index.

In summary, government geographic datasets, publicly accessible remote sensing products, and meteorological observations made within the most current time frame available for each dataset were used to compile all conditioning factors. Prior to the modeling stage, all vector-based layers were converted to raster format. The continuous variables were examined for missing values and spatial consistency, and the categorical variables (e.g., lithology, land use, and soil type) were reclassified into ranked susceptibility categories based on their observed relationship with hazard occurrence.

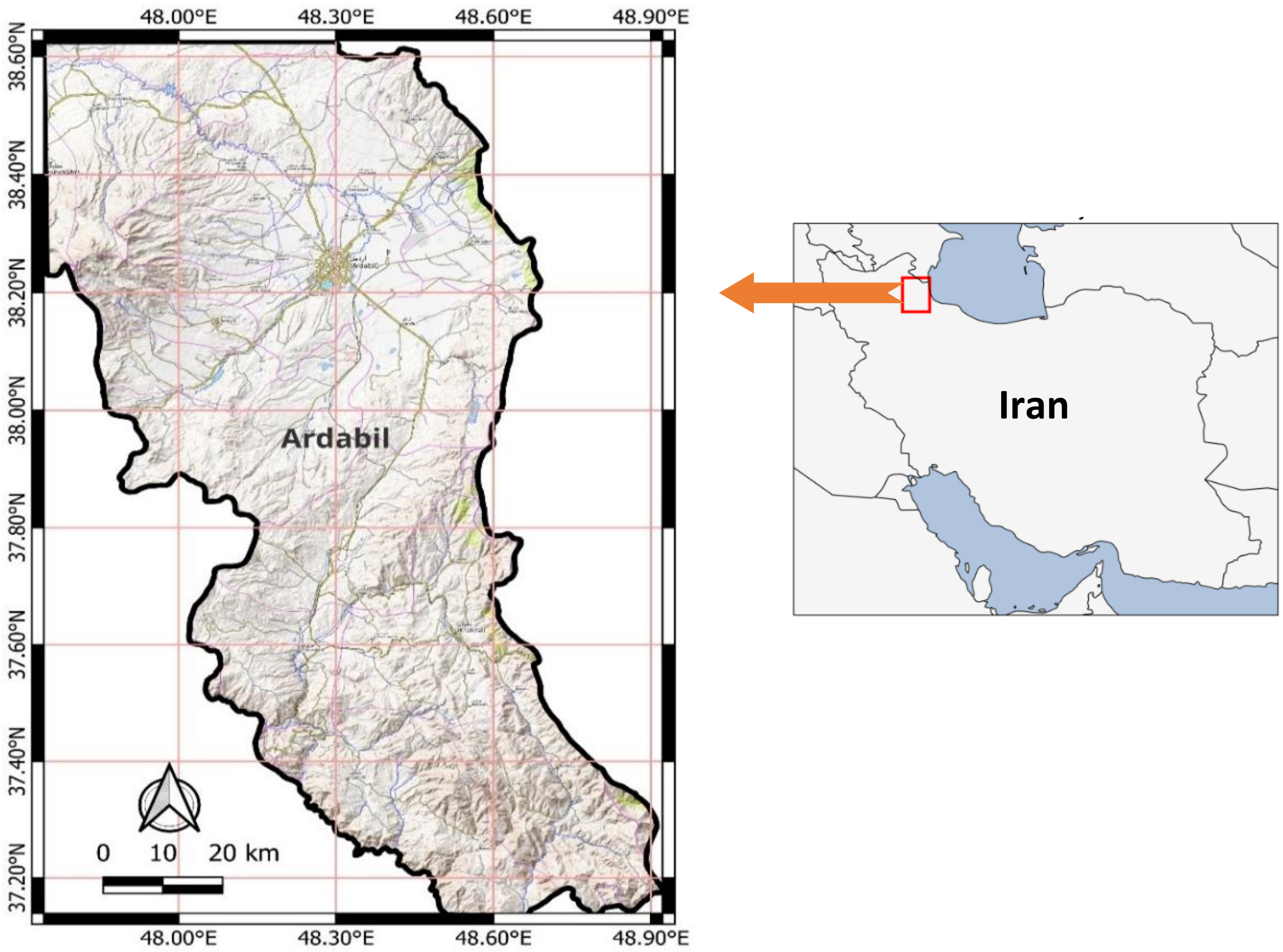

Figure 2. Location of Ardabil study area in Iran

A common valid-data mask was employed to eliminate missing or incorrect pixels from the study. Finally, all layers were transformed into a unified coordinate system (UTM), standardized to a common study-area extent, and resampled to a 30 × 30 m spatial resolution to enable pixel-based analysis and integrated spatial modeling. Furthermore, 191 landslide inventory samples from the region, as provided by the Geological Organization, were used for the testing and training of the model. The prepared maps were standardized and formatted as raster inputs for the machine learning models—specifically for variations of the proposed SAGE-XGBoost framework implemented in Python, as demonstrated in **Figure 3**. As main aim of this research is to evaluate the SAGE-XGBoost framework and compare it with well-established and powerful machine learning methods in natural hazards susceptibility mapping, only the key preprocessing steps required for reproducibility are summarized here, while detailed GIS implementation procedures are beyond the scope of the present work.

### 3.2. Case Study of Wildfire Susceptibility Mapping (WFSM)

The Arasbaran forest region is located in the northern part of East Azerbaijan Province, along the border between Iran and the Republic of Azerbaijan, and extends approximately between 38.30° to 39.42° N latitude and 46.00° to 47.60° E longitude (**Figure 4**). This region constitutes a segment of the Hyrcanian–Caucasian floristic zone, distinguished by its mountainous topography, characterized by steep slopes, deep valleys, and highly variable elevations ranging from approximately 250 meters to over 2,200 meters above the sea level. The region's climate is classified as semi-humid to mountainous Mediterranean, with relatively warm and dry summers. The region is characterized by dense vegetation, comprising species such as oak, hornbeam, maple, beech, and a variety of shrub species.

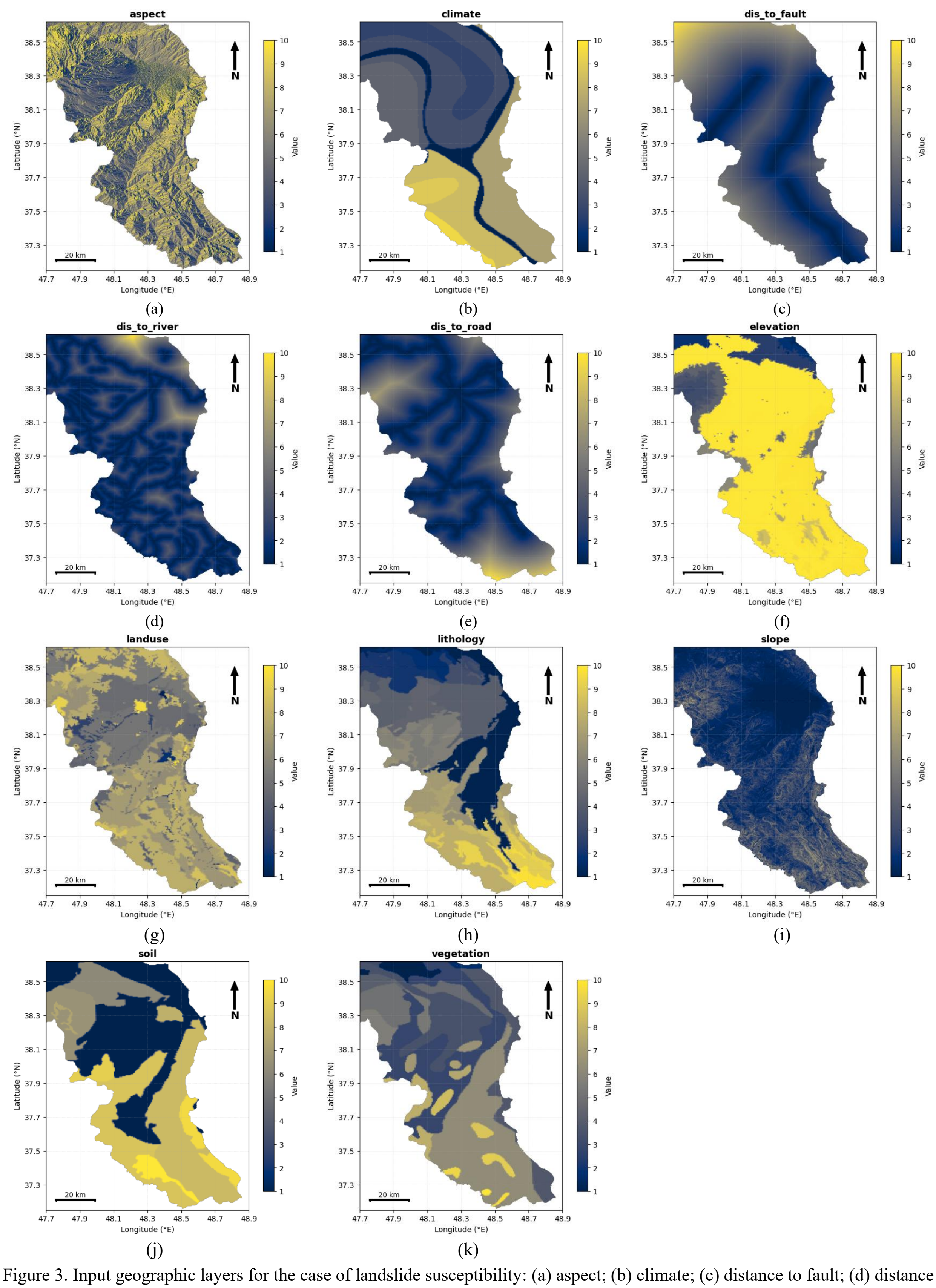

Figure 3. Input geographic layers for the case of landslide susceptibility: (a) aspect; (b) climate; (c) distance to fault; (d) distance to river; (e) distance to road; (f) elevation; (g) landuse; (h) lithology; (i) slope; (j) soil; (k) vegetation

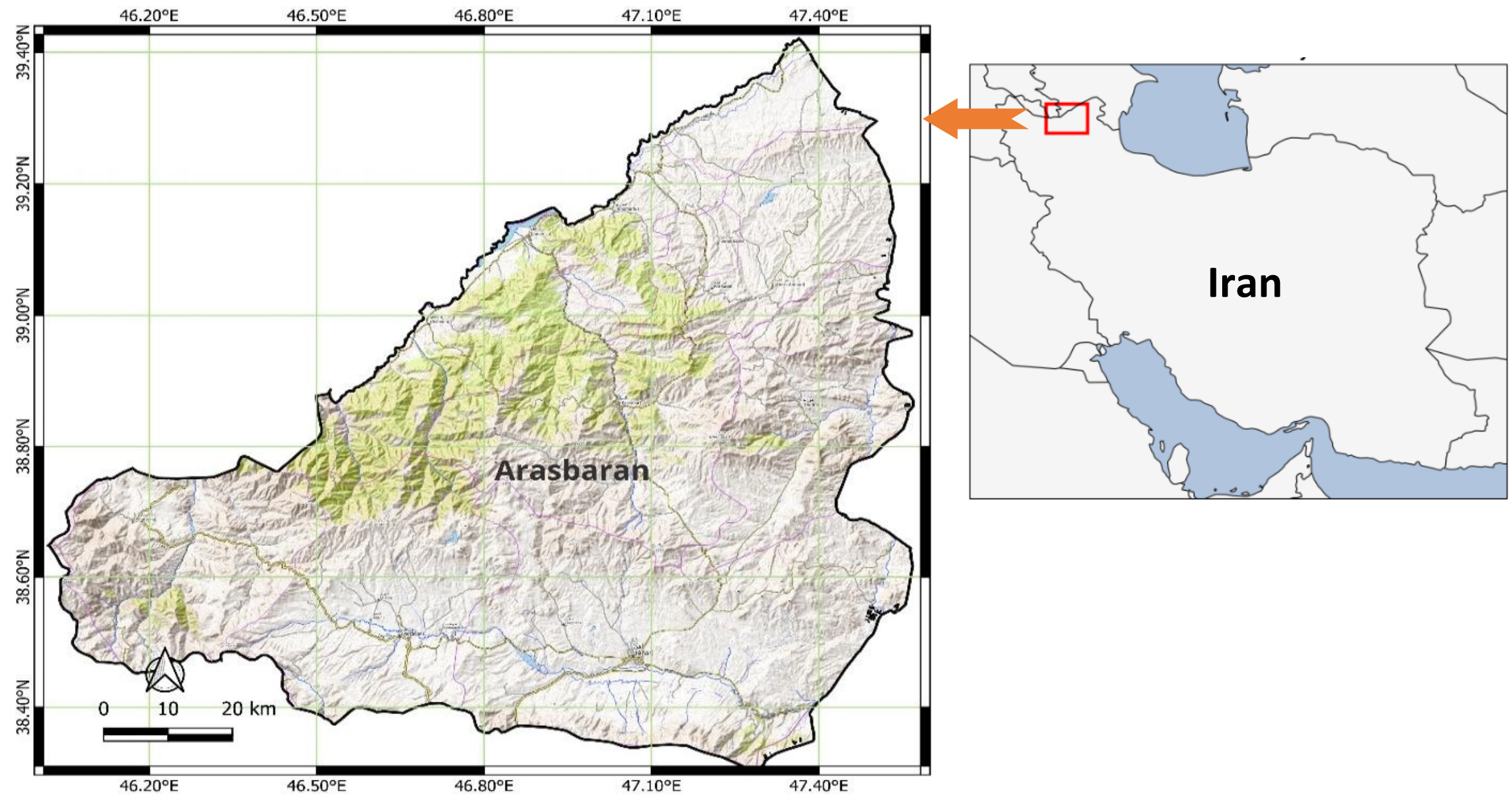

Figure 4. Location of Arasbaran study area in Iran

The area's susceptibility to forest fires is attributed to several factors, including high vegetation density, the presence of surface fuels such as leaf litter and woody debris, steep slopes, local wind systems, and rising temperatures during the warm season. Furthermore, anthropogenic pressures, including livestock grazing, tourism, and local resource exploitation, have the potential to exacerbate fire risk. These characteristics render Arasbaran a suitable region for forest fire susceptibility assessment and hazard zonation studies.

In the forest fire susceptibility assessment of the Arasbaran region, fourteen criterion layers were prepared and integrated as model input variables, including aspect, distance to road, elevation, humidity, land use, NDVI, plan curvature, precipitation, solar radiation, slope, stream network, temperature, Topographic Wetness Index (TWI), and wind.

DEM was derived from SRTM data with a spatial resolution of 30 meters (https://earthexplorer.usgs.gov/). Morphometric analyses within the GIS environment were employed to extract slope, elevation, aspect, plan curvature, and TWI layers. Climatic variables, including solar radiation, temperature, precipitation, and wind speed, were attained from global climate products and reanalysis datasets, such as WorldClim and CHELSA (https://chelsa-climate.org/), at appropriate spatial resolutions. These datasets were available as raster layers representing long-term or annual means. They were clipped to the study area and resampled into a 30-meter spatial resolution prior to analysis.

The derivation of land-use and NDVI layers was accomplished through the utilization of Landsat 8 OLI imagery (https://earthexplorer.usgs.gov/). Road and stream network data were obtained from available road authority maps and regional water management organizations (https://frw.ir/), and the distance-to-road layer was produced through the technique of Euclidean Distance analysis. Finally, all layers were projected into a unified UTM coordinate system, clipped to the common study-area extent, and resampled to a consistent 30 × 30 m spatial resolution. The prepared maps were standardized and formatted as raster inputs for the machine learning models—specifically for variations of the proposed framework implemented in Python, as illustrated in **Figure 5**. Furthermore, 128 wildfire inventory samples from the region, as provided by the Forest and Rangelands Organization, were gathered for the testing and training of the model.

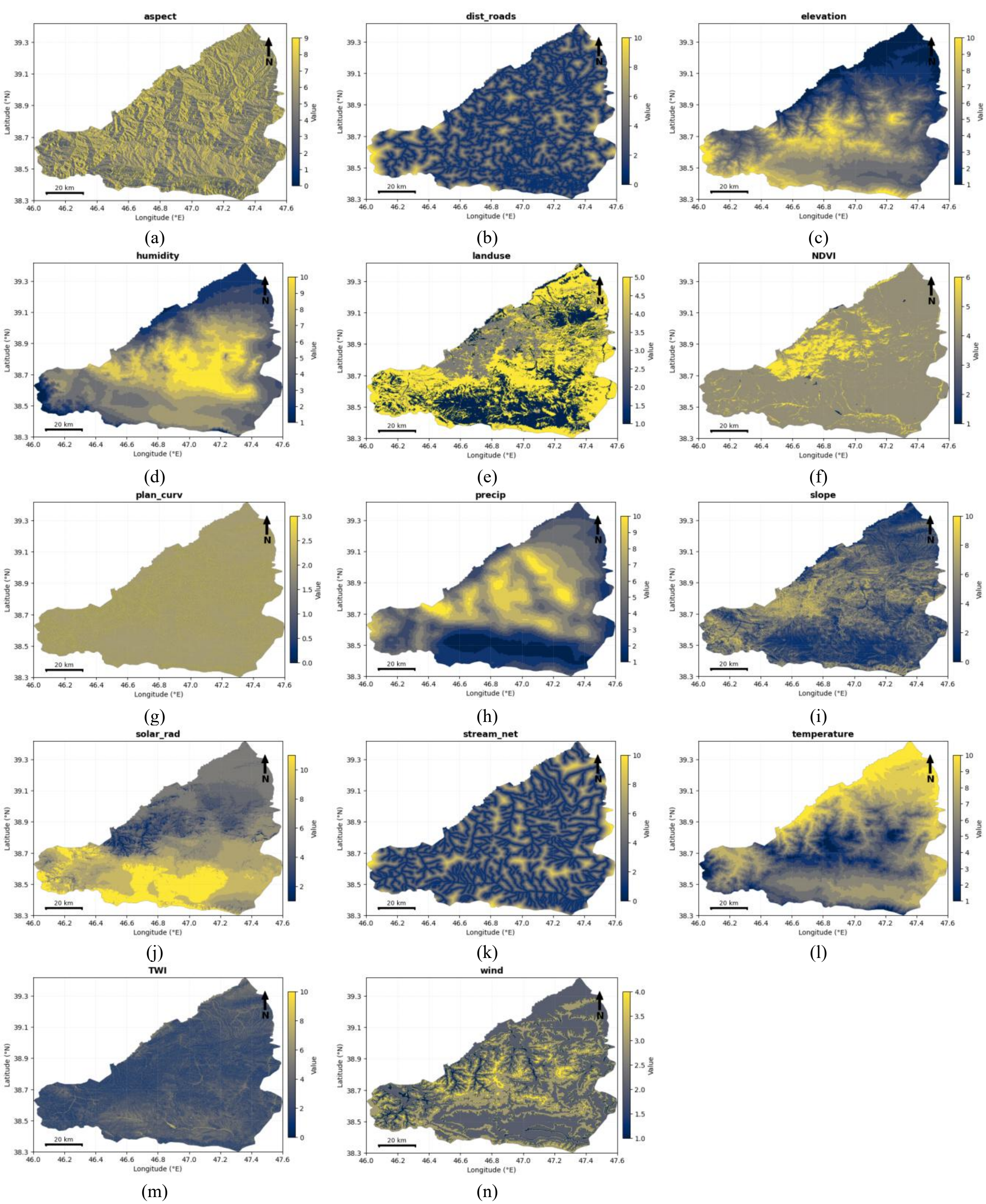

Figure 5. Input geographic layers for the case of wildfire susceptibility: (a) aspect; (b) distance to road; (c) elevation; (d) humidity; (e) landuse; (f) NDVI; (g) plan curvature; (h) precipitation; (i) slope; (j) solar radiation; (k) stream network; (l) temperature; (m) TWI; (n) wind

# 4. Results

## 4.1. Hyperparameter Settings and Generated Features from SAGE

In this section, three baseline machine learning algorithms—Random Forest (RF), Extreme Gradient Boosting (XGBoost), and Support Vector Machine (SVM)—along with two spatially explicit machine learning algorithms, Geographically Weighted Random Forest (GWRF) and Spatial XGBoost (S-XGBoost), were employed as the foundation for the initial analyses. In this context, S-XGBoost refers to the conventional XGBoost model augmented with spatial coordinates as additional input features. No explicit spatial weighting function, graph construction, local model fitting, or geographically weighted learning mechanism was employed. Therefore, S-XGBoost functions as a simple coordinate-enhanced baseline for evaluating the added values of the proposed SAGE-XGBoost framework. SAGE-XGBoost denotes the implementation of the proposed SAGE framework on XGBoost, selected due to its strong performance among state-of-the-art ML methods. Specifically, in contrast to S-XGBoost, SAGE-XGBoost incorporates the proposed noise-based data augmentation strategy and graph embeddings, while also including spatial coordinates as input features. Consequently, SAGE-XGBoost augments S-XGBoost by incorporating spatial feature enhancement and graph-based representation learning within the SAGE framework.

For all methods, the dataset was divided into testing and training subsets using a stratified sampling strategy with a 70:30 ratio. To obtain optimal models, dedicated pipelines were defined for each algorithm, allowing multiple combinations of key hyperparameters to be evaluated. All feasible hyperparameter combinations were tested, and the best-performing configuration for each method was determined based on Precision, Accuracy, F1-Score, and Recall. For the RF model, the hyperparameters included max_depth = 5, 10, 20, and None; n_estimators = 100, 200, 300, and 400; and min_samples_split = 2 and 5. For SVM, the regularization parameter C was tested with values of 1, 10, and 20, using both radial basis function (RBF) and linear kernels. For XGBoost and S-XGBoost, the evaluated hyperparameters included max_depth = 3, 4, and 6; n_estimators = 100, 200, 300, and 400; learning_rate = 0.05 and 0.1; and subsample = 0.8 and 1.0. In addition, to ensure robustness and mitigate overfitting, a five-fold cross-validation technique was employed during the training phase.

Spatial weighting was incorporated into the geographically weighted random forest (GWRF) model, leveraging a Gaussian kernel function to calculate the weights: $w_i = exp(-d_i^2/2h^2)$, where $h$ denotes the bandwidth parameter and $d_i$ represents the Euclidean distance between a training sample and the prediction location. The selection of the optimal bandwidth value was conducted by employing three distinct values: 20, 40, and 60. All training samples were utilized to construct a local random forest model for each prediction location. In addition, the Gaussian distance-decay weights were employed to regulate the influence of the samples, while the number of trees (100 and 200) and the maximum tree depth (5 and 10) were also optimized using weighted F1-score.

The graph embedding procedure, as represented in **Figure 6**, substantially increased the contextual information available for each sample prior to dimensionality reduction. In the landslide case, 11 environmental variables were initially utilized. For each variable and each sample, three neighborhood-based statistics—Neighborhood Mean, Neighborhood Standard Deviation, and Distance-Weighted Mean—were computed, resulting in 3 × 11 = 33 graph-derived features. Furthermore, the calculation of one Local Spatial Density feature resulted in a total of 34 graph-based features per sample. Principal Component Analysis (PCA) was subsequently implemented to diminish the 34 features to the assumed three, compact embedding components. The rationale behind the selection of three retained components from PCA as the ultimate output of graph embeddings was that it exhibited the most optimal performance, albeit with a marginal disparity, within the range of 2 to 5. Subsequent to the integration of these embeddings with the original environmental variables (11 features) and two spatial coordinate features (row and column), the final input space consisted of 11 (environmental) + 2 (spatial) + 3 (embedding), resulting in a total of 16 features.

In the wildfire case, 14 environmental variables were utilized. The same graph-based feature extraction strategy yielded 3 x 14 = 42 neighborhood statistics, 1 Local Spatial Density feature, and 43 graph-based features. Following the implementation of PCA reduction to three embedding components in conjunction with the integration of the original environmental and spatial features, the final feature space was determined to be 14 (environmental) + 2 (spatial) + 3 (embedding), resulting in a total of 19 features.

As represented in **Table** 1, a restricted set of candidate values was evaluated for the main SAGE-XGBoost hyperparameters. Specifically, the neighborhood size K was tested using {5, 10, 15, 20}, the PCA embedding dimension using {2, 3, 4, 5}, and the augmentation rate using several candidate values between 0.1 and 1.9.

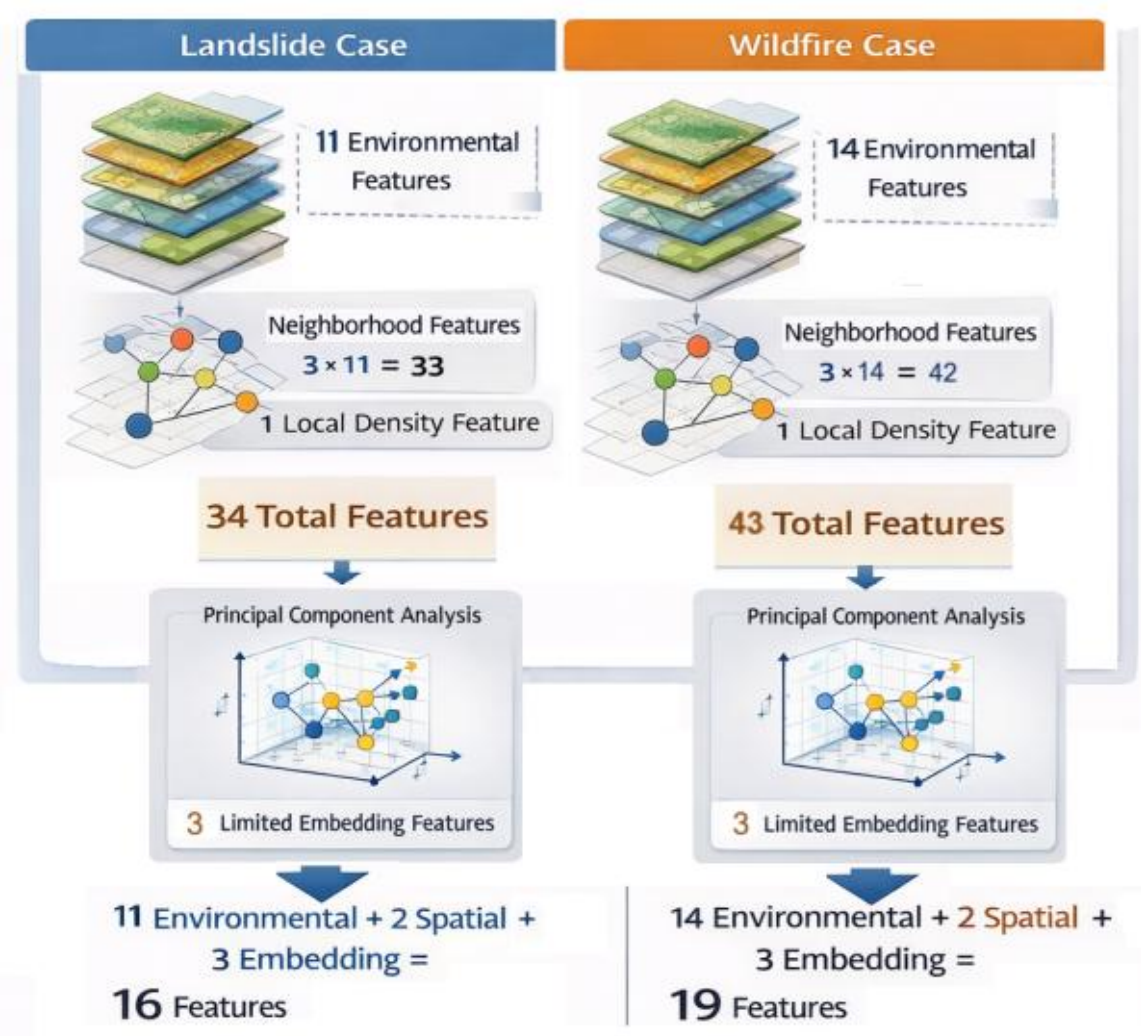

Figure 6. Determining the number of input features in each of landslide and wildfire cases based on the SAGE-XGBoost model

Table 1. Main hyperparameter sensitivity in SAGE-XGBoost

| Parameter | Tested Values | Selected |
|---|---|---|
| **K** | 5, 10, 15, 20 | 15 and 20 |
| **PCA Components** | 2, 3, 4, 5 | 3 |
| **Augmentation Rate** | 0.3, 0.5, 0.7, 0.9, 1.1, 1.3, 1.5 | 1.3 |

For each parameter combination, model performance was assessed using 5-fold cross-validation, and the optimal configuration was selected according to the mean weighted F1-score. Using this procedure, the configuration K = 15 (for landslide susceptibility) and K = 20 (for wildfire susceptibility), PCA dimension = 3, and augmentation rate = 1.3 consistently produced the best cross-validation performance and was therefore adopted in the final framework.

**Table 2** presents the inventory size and event density for both case studies, in order to provide a more quantitative description of data scarcity. The landslide inventory contains a mere 191 documented events, while the wildfire inventory encompasses 128 events distributed across extensive regional-scale study areas. When evaluated in relation to the spatial extent of the model, this suggests low event densities of 20 per 1,000 km² in landslide case study and 13.9 per 1,000 km² in wildfire case study. Subsequent to the subdivision of the data into four susceptibility classes, the number of training samples available per class becomes considerably smaller, thereby increasing the difficulty of learning robust class boundaries.

Table 2. Area of case studies and event density evaluation

| Dataset | Study Area (km²) | Initial Inventory Events | Event Density (events/km²) | Classes of Hazards |
|---|---|---|---|---|
| **Landslide** | **9,538** | 191 | 0.0200 | 3 |
| **Wildfire** | **9,198** | 128 | 0.0139 | 3 |

Consequently, the challenge addressed in this study is not merely limited sample count, but also a lack of spatial coverage and an inadequate depiction of environmental conditions throughout the research region.

The observed intensity or severity level of each recorded event was described by an attribute in the event inventory databases. The initial inventory source provided these severity levels, which were then utilized to establish three event classes (Classes 1-3) that corresponded to progressively higher impact levels. The final four-class susceptibility dataset was created by sampling non-event sites to add a fourth class (Class 0). Consequently, the proposed framework employed the classifications from the original inventory database instead of creating or modifying the event severity labels.

To mitigate potential class imbalance, non-event samples were selected using a controlled sampling strategy rather than random extraction from the entire study area. The number of background samples was maintained within the range of the event-class sample counts, resulting in relatively balanced class distributions across the four susceptibility classes. A summarization of data distribution in each class by accounting for data augmentation is presented in **Table 3**. For the landslide susceptibility dataset, the test samples in classes 0–3 were 30, 29, 51, and 29, respectively. For the wildfire susceptibility dataset, the corresponding sample counts were 24, 36, 17, and 21. These distributions indicate a moderate variation among classes rather than a severe imbalance. Consequently, macro-averaged and weighted-averaged F1-scores produced nearly identical values, indicating that class imbalance had a negligible effect on model evaluation.

## 4.2. Acquired Performance Metrics

The results in this experiment are illustrated in **Figure 7**. The proposed SAGE-XGBoost framework demonstrates a substantial improvement in performance across all evaluation metrics when compared to extant baseline models. The accuracy of the model increased from 0.55 (S-XGBoost) to 0.88, representing a relative improvement of approximately 60%.

Table 3. Distribution of samples in each category of hazards

| Dataset | Resolution of Predictor Layers (m) | | Class 0 (Low) | Class 1 (Moderate) | Class 2 (High) | Class 3 (Very High) |
|---|---|---|---|---|---|---|
| **Landslide** | **30** | **Total Samples** | 100 | 97 | 169 | 95 |
| | | **Test Samples** | 30 | 29 | 51 | 29 |
| **Wildfire** | **30** | **Total Samples** | 78 | 119 | 56 | 76 |
| | | **Test Samples** | 24 | 36 | 17 | 21 |

A similar enhancement was observed in the F1-score, which increased from 0.53 to 0.87, suggesting a substantial improvement in balanced classification performance. The performance of traditional ML models (SVM, RF, and XGBoost) is moderate, with accuracy levels varying between 0.48 and 0.53. Incorporating spatial coordinates (S-XGBoost) results in marginal improvement. Conversely, the incorporation of data augmentation and graph embeddings in SAGE-XGBoost results in a substantial enhancement in performance, underscoring the efficacy of spatial contextual learning.

In the context of wildfire susceptibility assessment, SAGE-XGBoost once again exhibits superior performance, attaining an accuracy of 0.83, in comparison to 0.55 for S-XGBoost and 0.59 for GWRF (the strongest baseline). The F1-score demonstrates a notable enhancement, progressing from 0.50 (S-XGBoost) to 0.82, signifying an advancement in precision–recall equilibrium. Conventional XGBoost demonstrates comparatively weak performance (Accuracy = 0.47), indicating that wildfire susceptibility is more sensitive to spatial context modeling. The consistent improvement observed in both case studies lends credence to the robustness and general applicability of the proposed SAGE framework.

### 4.3. Confusion Matrix Analysis

**Figure 8** presents a heatmap-style visualization of the normalized confusion matrices for the landslide susceptibility assessment, illustrating the predictive performance across different classes (low, moderate, high, and very high) separately for each method. In these visualizations, improved model performance is reflected by brighter colors (yellow to light green) along the main diagonal and darker tones in the off-diagonal elements, indicating higher correct classification rates and reduced misclassification, respectively.

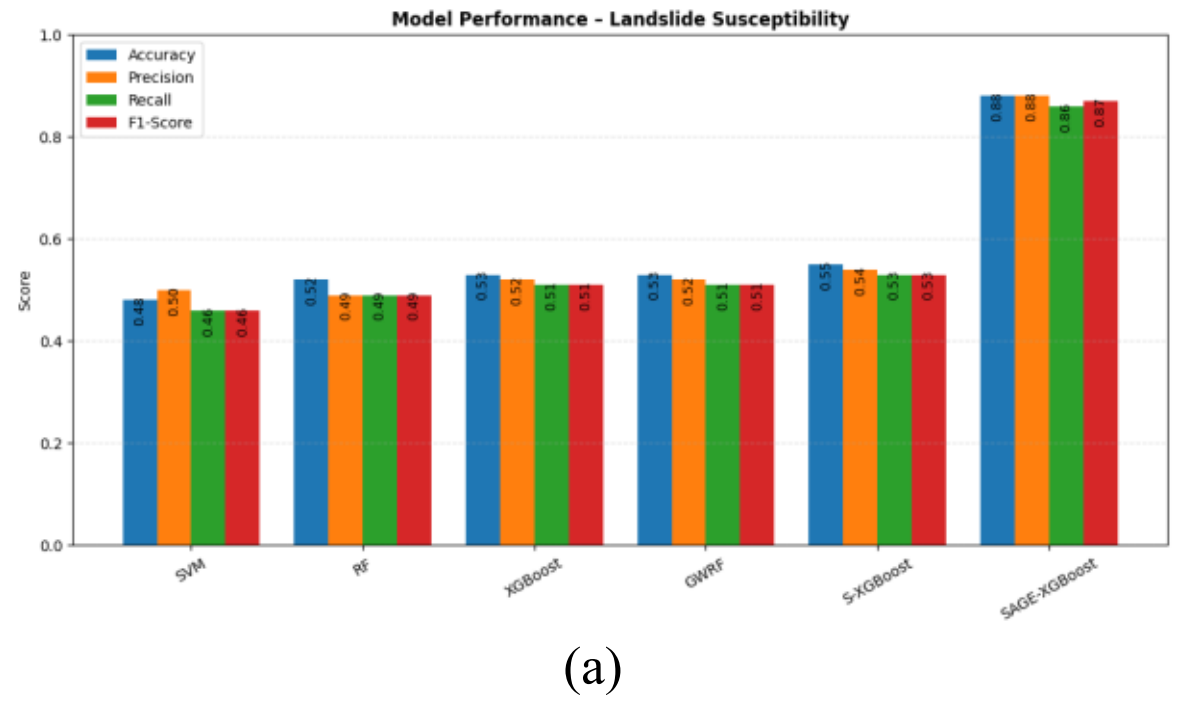

(a)

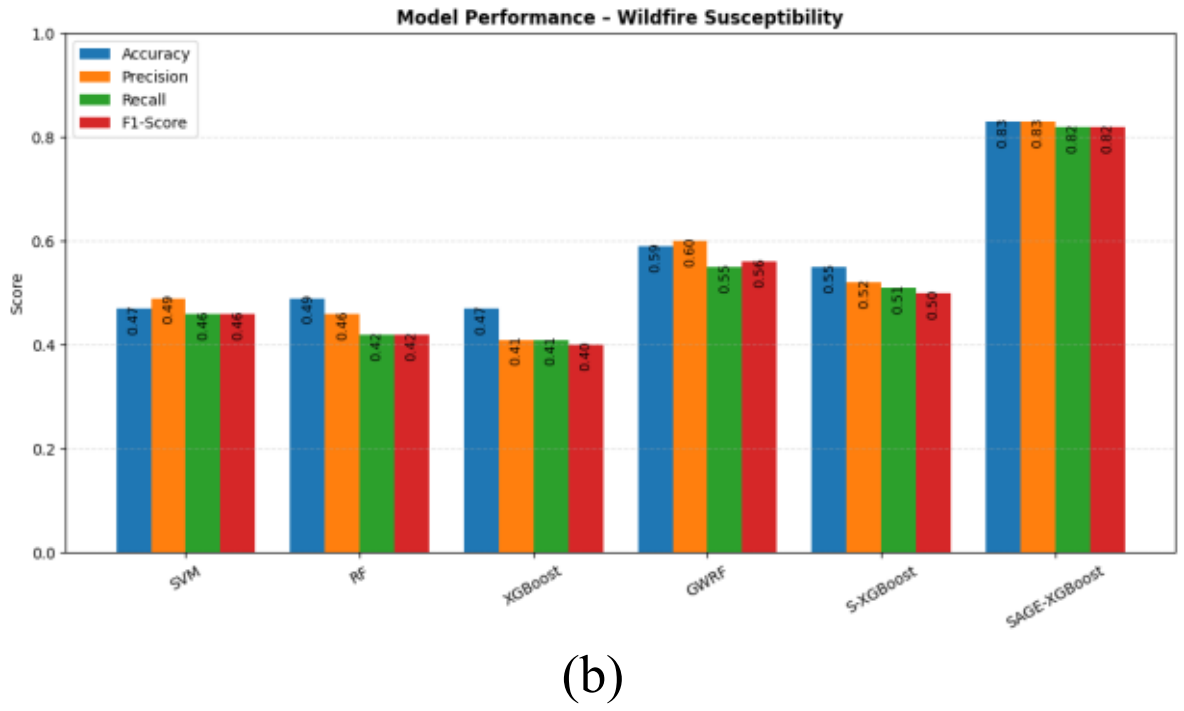

(b)

Figure 7. Comparison of performance metrics in six method of SVM, RF, XGBoost, GWRF, S-XGBoost, and SAGE-XGBoost

The findings indicate that SAGE-XGBoost exhibits the most robust performance across all four susceptibility classes, with main diagonal values of 96.7%, 72.4%, 96.1%, and 86.2%, respectively. These results suggest that the model exhibits high precision, particularly within the low and high susceptibility classes. In the subsequent analysis, the Geographical Random Forest (GWRF) algorithm exhibited a commendable performance, albeit with a substantial margin of difference. It attained diagonal values of 68.8%, 50.0%, 73.1%, and 46.7% for the four distinct classes.

Compared to the low and high susceptibility classes, the moderate susceptibility class exhibited a lower classification accuracy. Since the moderate class is a transitional zone between adjacent susceptibility levels, this behavior is to be expected in multiclass susceptibility mapping. Hence, samples in the moderate class often share environmental traits with both the lower- and higher-susceptibility classes. Consequently, there is more feature space overlapping and classification ambiguity. The low and high susceptibility classifications, on the other hand, are usually easy to distinguish.

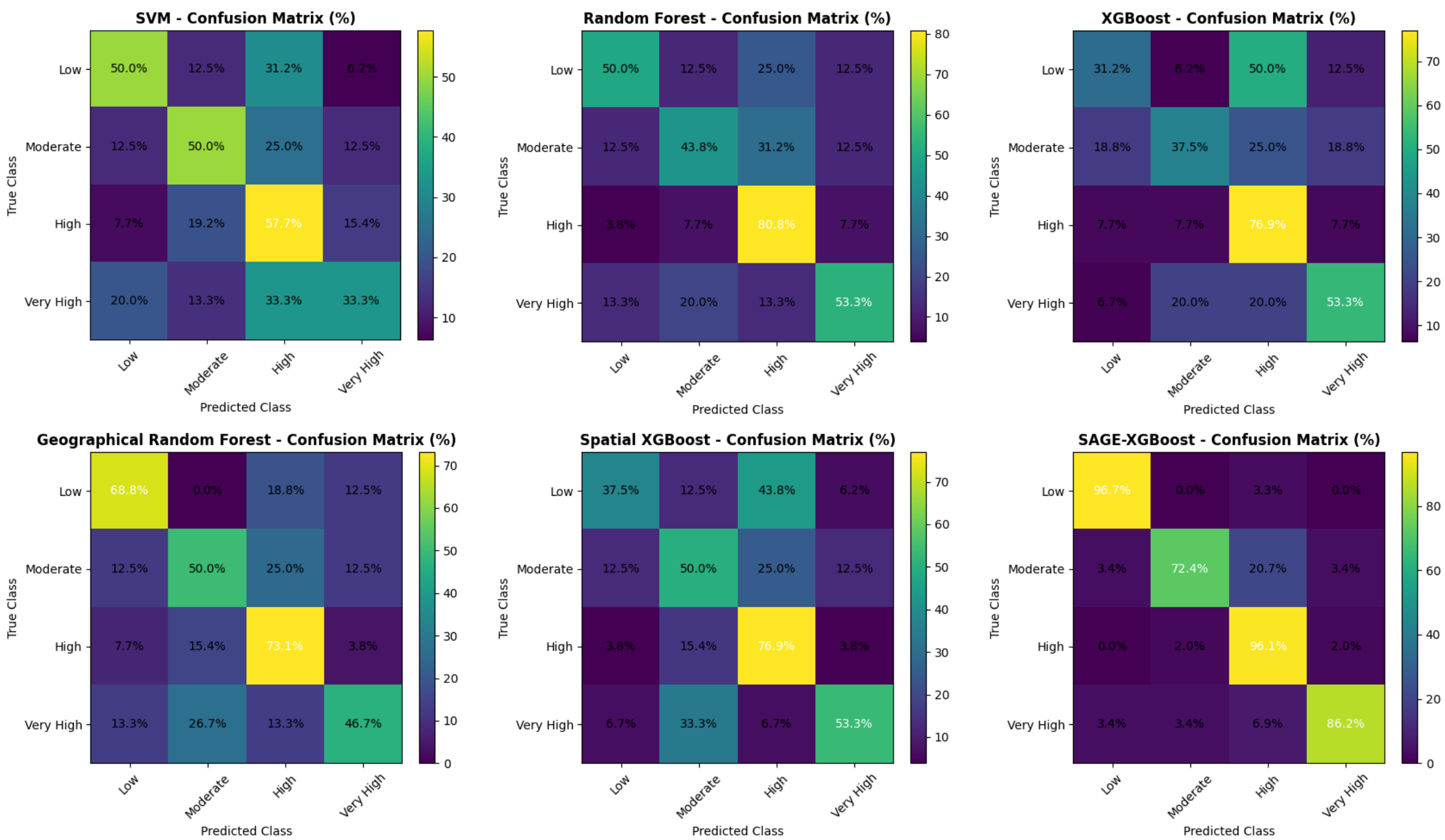

Figure 8. Heatmap-style visualization of confusion matrix to evaluate predictions in landslide susceptibility categories, separately for each method

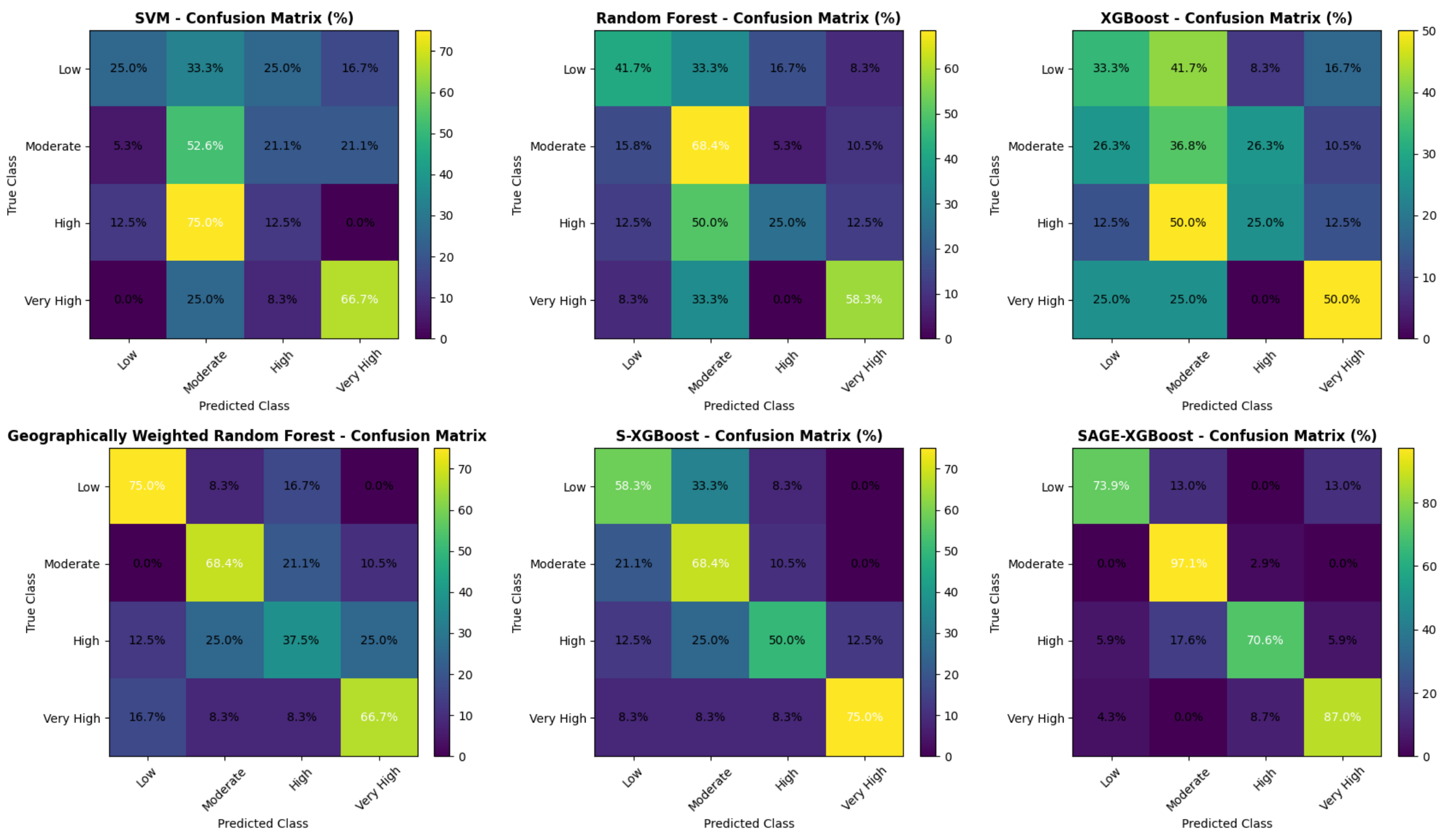

Figure 9. Heatmap-style visualization of confusion matrix to evaluate predictions in wildfire susceptibility categories, separately for each method

A comparable heatmap-style confusion matrix visualization for the wildfire susceptibility case is presented in **Figure 9**. Once more, SAGE-XGBoost demonstrates superiority over all competing methods, attaining diagonal values of 73.9%, 97.1%, 70.6%, and 87.1% for the low, moderate, high, and very high classes, respectively. The second-best performance in this case is obtained by Spatial XGBoost (S-XGBoost), with diagonal values of 58.3%, 68.4%, 50.0%, and 75.0%, indicating that incorporating spatial coordinates alone improves performance, but remains substantially inferior to the full SAGE framework.

### 4.4. ROC Curve and AUC Comparison

A one-versus-rest (OvR) multi-class framework was employed to analyze ROC-AUC, given that susceptibility mapping was initially developed as a four-class classification problem. A distinct ROC curve was created for each class based on the expected probability after class labels were first binarized. The reported area under the curve (AUC) values correspond to the macro-average ROC-AUC, where all classes contribute equally, regardless of class frequency. This approach was adopted to ensure an equitable evaluation of discrimination performance at all levels of susceptibility.

**Figure 10** demonstrates the Receiver Operating Characteristic (ROC) curves for the applied machine learning models in the landslide (**Figure 10-a**) and wildfire (**Figure 10-b**) case studies. The matching Area Under the Curve (AUC) values, provided in the legend, serve as quantitative indicators for method comparison.

In both case studies, SAGE-XGBoost achieves the highest AUC values, reaching 0.975 in the landslide study and 0.951 in the wildfire study. These values indicate near-optimal separability between susceptibility classes and confirm the superior discriminative capability of the proposed framework compared to conventional and spatially explicit baseline models including GWRF, and Spatial XGBoost.

## 5. Discussions

### 5.1. Cross-Region Transferability

Although the models were initially evaluated using conventional random 5-fold cross-validation, random partitioning may ignore spatial autocorrelation and can lead to information leakage when spatially proximate samples are assigned to both training and testing subsets (Roberts et al., 2017). To address this issue, an additional spatial cross-validation experiment was conducted. The study area was divided into a 5 × 5 spatial grid, and non-empty blocks were used as groups for block cross-validation. In each fold, one or more spatial blocks were reserved for testing, while the remaining blocks were used for training. This approach evaluates model performance in spatially unseen regions and provides a more rigorous assessment of spatial generalization.

The results for XGBoost, Spatial XGBoost (S-XGBoost), and the proposed SAGE-XGBoost are reported in **Table 4** and **Table 5**. As expected, both accuracy and weighted F1-score decreased under spatial cross-validation compared with conventional random validation, reflecting the increased difficulty of predicting susceptibility in unseen spatial domains. Nevertheless, SAGE-XGBoost consistently achieved the best performance. For landslide susceptibility mapping, it obtained an accuracy of 0.60 and a weighted F1-score of 0.59, compared with 0.47 and 0.46 for S-XGBoost, while XGBoost produced the lowest scores. Similarly, for wildfire susceptibility mapping, SAGE-XGBoost achieved an accuracy of 0.56 and a weighted F1-score of 0.54, outperforming S-XGBoost (0.37 accuracy and 0.34 weighted F1-score) and XGBoost.

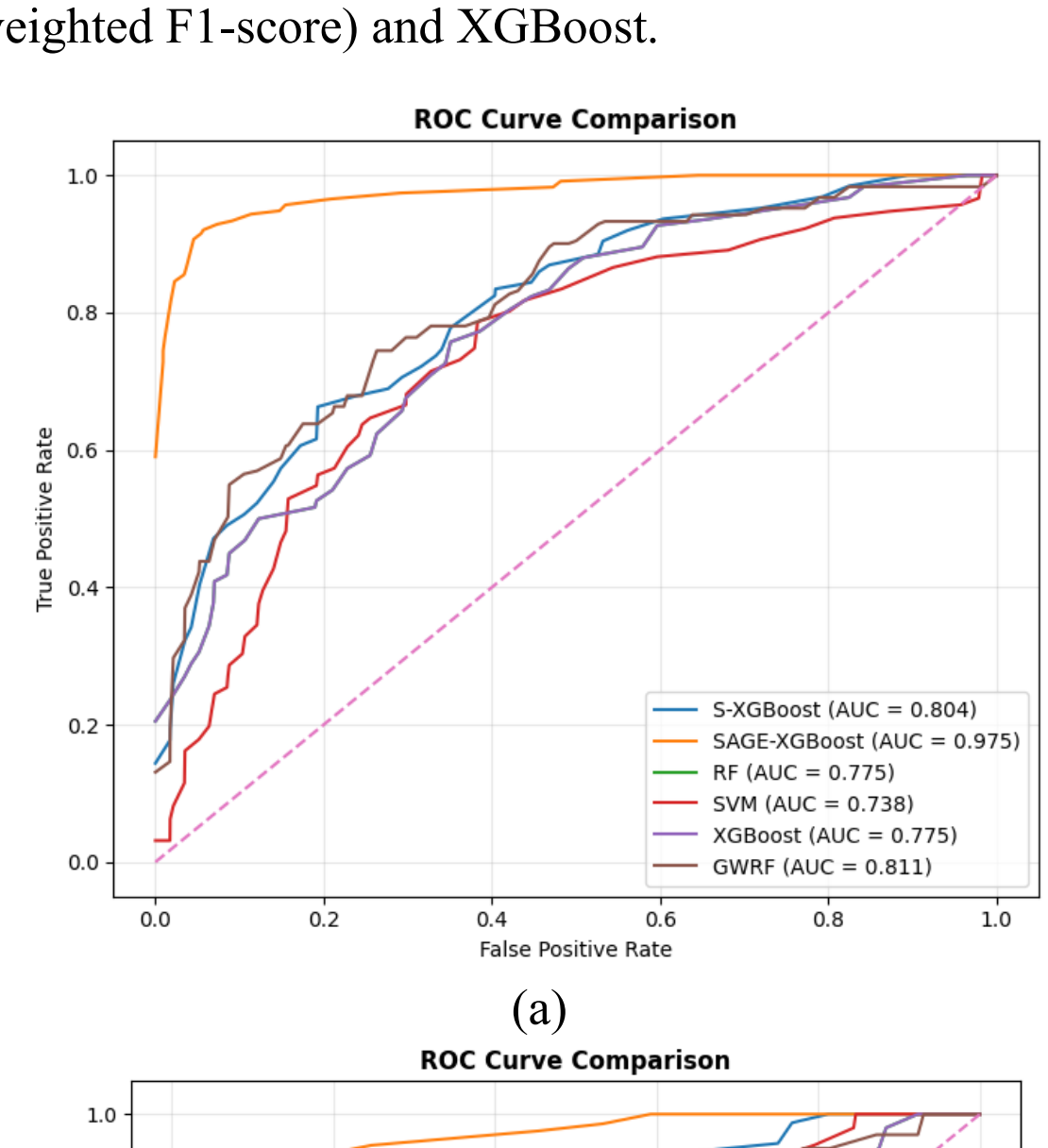

(a)

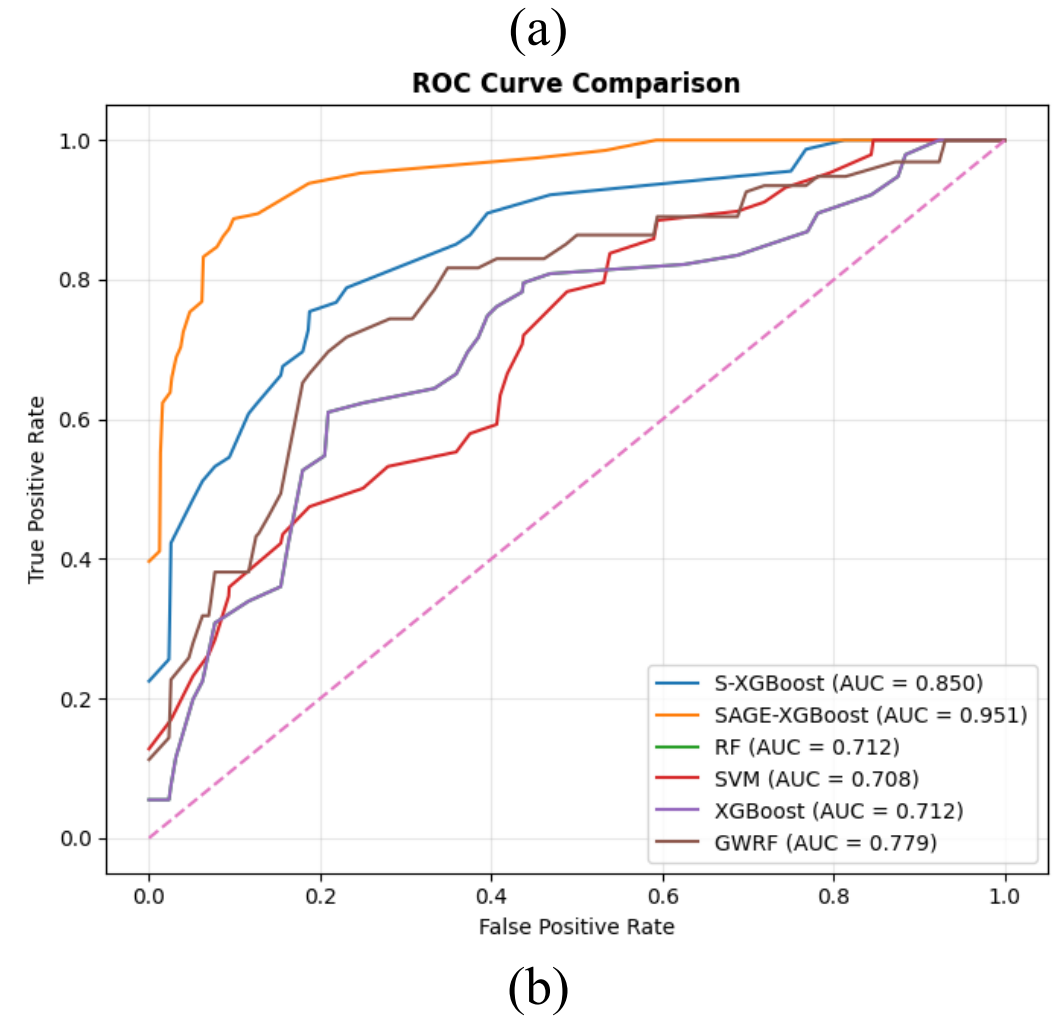

(b)

Figure 10. Comparison of ROC-AUC of methods: (a) the case of landslide susceptibility; (b) the case of wildfire susceptibility

These results indicate that, despite the stricter evaluation imposed by spatial cross-validation, SAGE-XGBoost maintains a clear advantage over the baseline models. This suggests that the proposed graph-based feature engineering, neighborhood embedding, and data augmentation strategies contribute positively to spatial generalization. It should also be noted that the primary objective of SAGE-XGBoost is to improve susceptibility modeling under data-scarce conditions rather than maximizing transferability across completely unseen geographic regions. Further research may explore advanced spatial learning and domain-transfer strategies to improve cross-region generalization.

Table 4. Evaluation of methods based on block cross validation for the case of landslide susceptibility

| Method | Accuracy (Avg.) | F1-Score (Avg.) |
|---|---|---|
| **XGBoost** | 0.45 | 0.44 |
| **S-XGBoost** | 0.47 | 0.46 |
| **SAGE-XGBoost** | 0.60 | 0.59 |

Table 5. Evaluation of methods based on block cross validation for the case of wildfire susceptibility

| Method | Accuracy (Avg.) | F1-Score (Avg.) |
|---|---|---|
| **XGBoost** | 0.34 | 0.33 |
| **S-XGBoost** | 0.37 | 0.34 |
| **SAGE-XGBoost** | 0.56 | 0.54 |

## 5.2. Technical Interpretation of Performance Improvements

SAGE-XGBoost's consistent superior performance across confusion matrix metrics and AUC analysis suggests that the proposed framework effectively enhances class separability under conditions of limited labeled data. This improvement is due to three complementary mechanisms:

(1) Controlled Data Augmentation: This strategy uses noise to increase the diversity of the sample while preserving the statistical structure of environmental variables. This mitigates the adverse effects of data scarcity and reduces the risk of underfitting, a common issue in conventional machine learning models trained on limited inventories.

(2) Graph-Based Contextual Feature Extraction: The model captures higher-order spatial dependencies and environmental heterogeneity by incorporating neighborhood statistics (mean, standard deviation, and distance-weighted mean) in conjunction with local spatial density. These graph-derived features implicitly encode spatial autocorrelation and contextual structure, both of which are usually overlooked in standard tabular machine learning approaches.

(3) Dimensionality-Controlled Embedding: The application of PCA to compress high-dimensional graph features into a compact embedding representation has been demonstrated to preserve structural information while preventing feature space inflation. This equilibrium between contextual enrichment and dimensionality reduction enhances generalization capability.

The proposed framework essentially offers an intermediate but efficient solution, whereas deep learning approaches necessitate significantly larger labeled datasets to prevent overfitting, and traditional machine learning models frequently suffer from underfitting in data-scarce environments. Significant performance improvements are achieved in both environmental case studies as a result of SAGE-XGBoost's transformation of the data scarcity problem into an opportunity for structured contextual learning by enriching information density per sample through spatial graph embeddings and controlled augmentation.

The proposed framework, consistent with the prevailing approach in susceptibility mapping studies, posits that inventory labels provide the most accurate representation of actual event conditions. Predictive performance may be impacted by any uncertainty related to inventory compilation or severity classification that spreads throughout the learning process. However, it should be noted that the same inventory sources and labels were consistently used for all competing models, ensuring a fair comparative evaluation.

### 5.3. Sensitivity Analysis of Augmentation Rate, PCA Embedding, and K Neighbors

In order to assess the impact of synthetic data generation, an analysis was conducted where the augmentation rate was modified from 0 (original data exclusively) to 1 (synthetic samples equivalent in quantity to the original dataset) with increments of 0.05. The performance of the model was evaluated by means of the classification error, which is defined as Error = 1 – Accuracy. The augmentation process was terminated when the marginal reduction in error between successive augmentation levels became smaller than 1%, ensuring that only statistically meaningful improvements were considered while preventing excessive synthetic sample generation.

In the landslide case, performance exhibited a consistent enhancement until an augmentation rate of 1.3, at which point the error reduction became negligible. A similar outcome was observed in the wildfire case, where stabilization occurred at an augmentation rate of 1.5. In both cases, the error curves manifest a discernible saturation behavior that extends beyond these established thresholds. This phenomenon can be elucidated through the lens of the bias-variance trade-off. In scenarios where the available data is labeled with a high degree of restriction, a strategy of controlled noise-based augmentation has been shown to yield notable benefits. This augmentation technique has been observed to increase variability within the intra-class category, thereby enriching the effective training distribution. This, in turn, has been demonstrated to lead to a reduction in bias and an enhancement in the estimation of decision boundaries. However, once sufficient variability has been introduced, additional synthetic samples contribute limited new information, leading to performance saturation. The identified optimal augmentation rates therefore reflect a balance between effective sample enrichment and redundancy control, supporting the robustness of the SAGE-XGBoost framework under data scarcity conditions.

In order to ascertain an appropriate graph embedding dimension, PCA configurations containing 2–5 principal components were evaluated using 5-fold cross-validation for both cases of landslide susceptibility and wildfire susceptibility. The cumulative explained variance and predictive performance for the first case were 71, 85, 89, and 92% for the dimension from 2 through 5, respectively. In the second case, these values were determined to be 67, 84, 86, and 90%, respectively. As illustrated in **Figure 11**, the relationship between embedding dimensionality, cumulative explained variance, and predictive performance is evident. The incorporation of additional principal components led to a consistent enhancement in the cumulative explained variance. However, beyond three components, the gains in accuracy and weighted F1-score became marginal. While four and five components exhibited slightly higher levels of variance, they did not result in significant performance enhancements, despite an increase in feature dimensionality and computational complexity. Consequently, three principal components were selected as the final embedding dimension, representing the optimal trade-off between information preservation, predictive performance, and model simplicity.

A thorough analysis of the PCA loading matrices was conducted, which revealed that the graph embedding dimensions that were retained correspond to environmental patterns that are meaningful at the neighborhood scale (**Table 6** and **Table 7**). In the landslide case, the dominant components were largely governed by fault-, road-, and river-related neighborhood statistics, whereas in the wildfire case the principal components were mainly controlled by temperature, elevation, humidity, precipitation, solar radiation, and aspect characteristics. These findings suggest that the proposed graph embedding does not merely represent abstract latent features; rather, it compresses physically interpretable spatial-neighborhood information into a compact feature space. Consequently, the PCA-based embedding preserves the most influential local environmental relationships while substantially reducing the dimensionality of the original graph-derived feature set.

A sensitivity study was carried out by changing the number of nearest neighbors (K) from 5 to 20 to assess the robustness of the suggested framework regarding neighborhood definition. Both accuracy and weighted F1-score showed a generally increasing trend with neighborhood size, as shown in **Figure 12**, and then stabilized within a relatively narrow range that included the optimum. K = 18 produced the maximum accuracy and weighted F1-score for wildfire susceptibility mapping, whereas K = 15 produced the best results for the landslide susceptibility dataset. The results show a relatively low sensitivity to the value of K, where reduced neighborhood sizes tended to yield marginally lower and less stable performance, presumably due to inadequate characterization of the local spatial context.

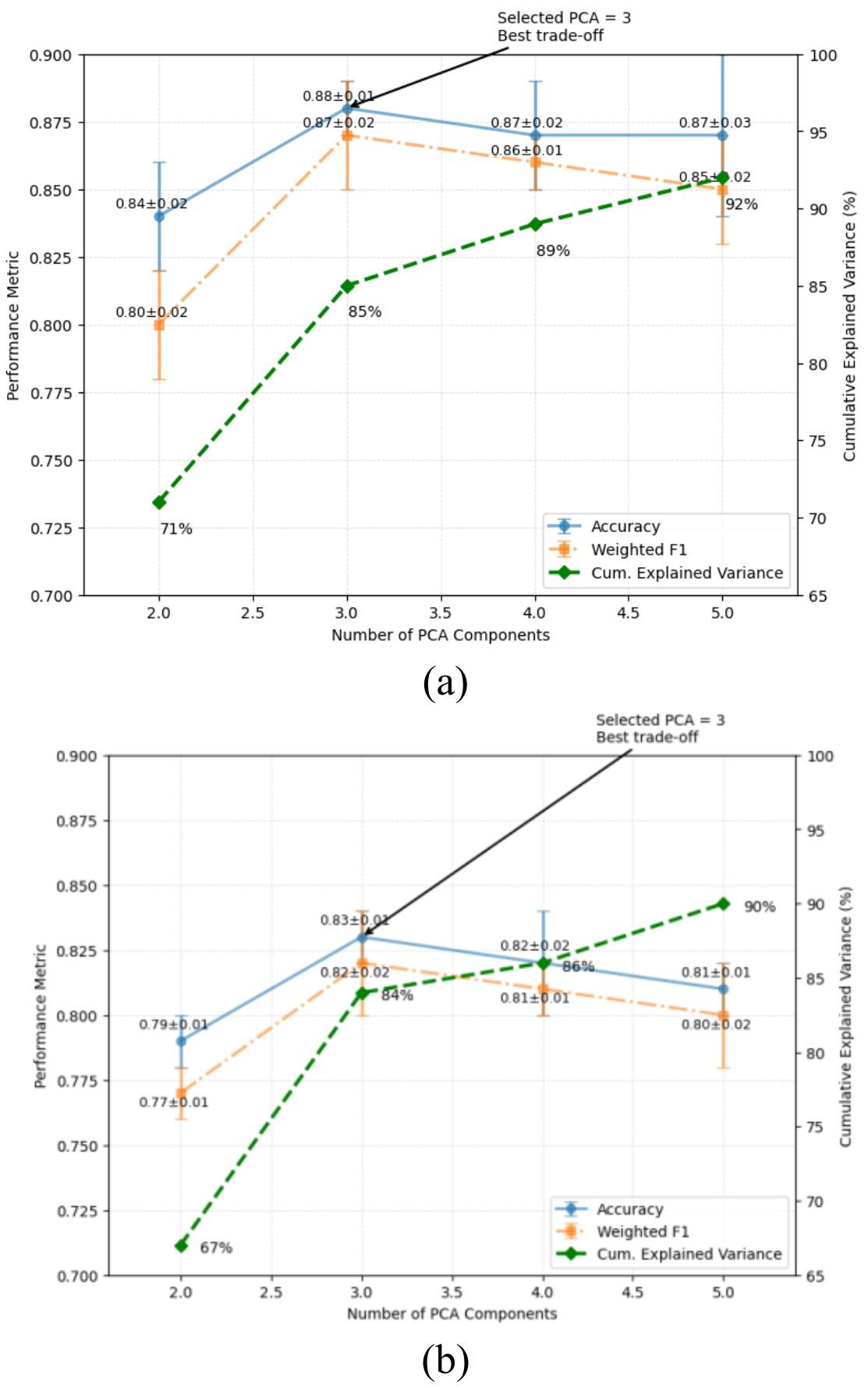

Figure 11. Selection of PCA embedding dimensions in SAGE-XGBoost: (a) the case of landslide susceptibility; (b) the case of wildfire susceptibility

Table 6. Top five contributors in each PCA component based on loading matrix in landslide susceptibility

| PC1 | | PC2 | | PC3 | |
|---|---|---|---|---|---|
| **Variable Name** | **Contribution** | **Variable Name** | **Contribution** | **Variable Name** | **Contribution** |
| Weighted Mean Distance to Fault | 0.61 | Weighted Mean Distance to Road | 0.62 | Weighted Mean Distance to River | 0.80 |
| Mean Distance to Fault | 0.53 | Mean Distance to Road | 0.53 | Mean Distance to River | 0.55 |
| Weighted Mean Distance to Road | 0.41 | Weighted Mean Distance to Fault | 0.41 | Standard Deviation Distance to River | 0.14 |
| Mean Distance to Road | 0.38 | Mean Distance to Fault | 0.38 | Standard Deviation Distance to Road | 0.08 |
| Standard Deviation Distance to Fault | 0.06 | Standard Deviation Distance to Fault | 0.04 | Mean Distance to Road | 0.07 |

Table 7. Top five contributors in each PCA component based on loading matrix in wildfire susceptibility

| PC1 | | PC2 | | PC3 | |
|---|---|---|---|---|---|
| **Variable Name** | **Contribution** | **Variable Name** | **Contribution** | **Variable Name** | **Contribution** |
| Weighted Mean Temperature | 0.44 | Weighted Mean Precipitation | 0.46 | Weighted Mean Aspect | 0.90 |
| Weighted Mean Elevation | 0.43 | Mean Precipitation | 0.42 | Mean Aspect | 0.32 |
| Mean Temperature | 0.37 | Weighted Mean Solar Radiation | 0.39 | Weighted Mean Stream Network | 0.15 |
| Mean Elevation | 0.36 | Mean Solar Radiation | 0.36 | Weighted Mean TWI | 0.09 |
| Weighted Mean Humidity | 0.29 | Weighted Mean Slope | 0.31 | Weighted Mean Slope | 0.08 |

## 5.4. Ablation Analysis and the Effect of Graph Embeddings Feature

At this stage of the evaluation, S-XGBoost was selected as the benchmark model, as it accounts for spatial heterogeneity and demonstrated slightly better performance than the other baseline machine learning models (excluding SAGE-XGBoost). In order to separately investigate the contribution of each component of the proposed framework, the data augmentation module was first applied using the optimal augmentation rates identified in the previous section for both the landslide and wildfire cases. This configuration enabled the isolation and quantification of the independent effects of the data augmentation strategy and the graph embedding features. The performance results are outlined in **Table 8** for the landslide case and **Table 9** for the wildfire case, with direct comparison to SAGE-XGBoost.

In the landslide susceptibility assessment, incorporating the graph embeddings feature enhanced the classification accuracy from 0.83 to 0.88, corresponding to an improvement of approximately 5.6%. Given that the baseline S-XGBoost model, devoid of data augmentation, attained an accuracy of 0.55 (as detailed in the Results section), the isolated contribution of the data augmentation component can be estimated at approximately 33.7%. In summary, the comprehensive SAGE-XGBoost framework results in an estimated overall enhancement of approximately 38% in comparison with the S-XGBoost model serving as the baseline. A similar pattern of relative gains is observed across other evaluation metrics, including precision, recall, and F1-score.

In the wildfire susceptibility case, the graph embedding features enhanced the accuracy from 0.78 to 0.83, signifying a substantial gain of approximately 6%. Given that the baseline S-XGBoost model, without augmentation, attained an accuracy of 0.55, the augmentation component is estimated to contribute approximately 29.4% to the overall accuracy. Consequently, the overall improvement achieved by SAGE-XGBoost in the wildfire case is approximately 33%. These findings suggest that while both components contribute to performance enhancement, the data augmentation module provides the dominant improvement under conditions of data scarcity, and the graph embedding features further refine spatial discrimination and class separability. The waterfall diagrams in **Figure 13** illustrate these advancements for the two case studies.

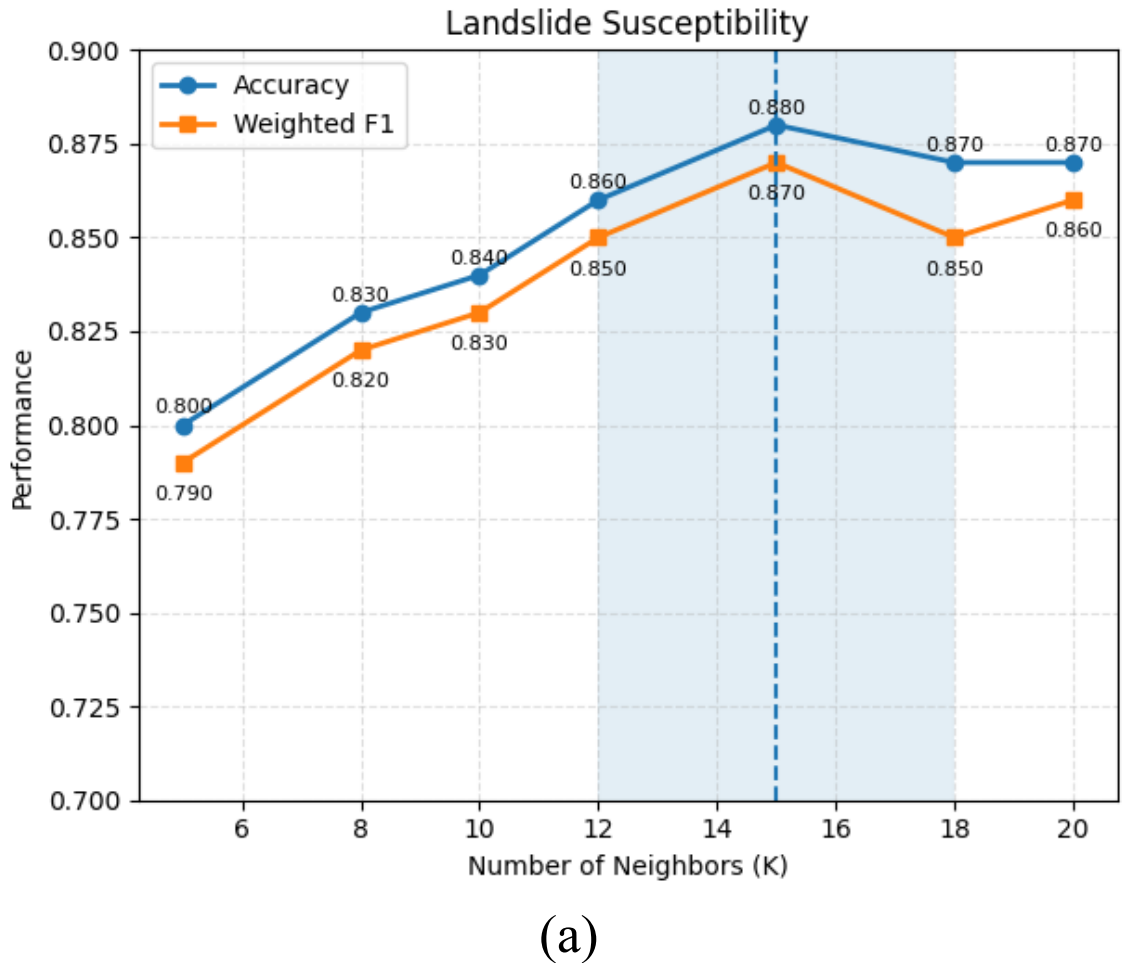

(a)

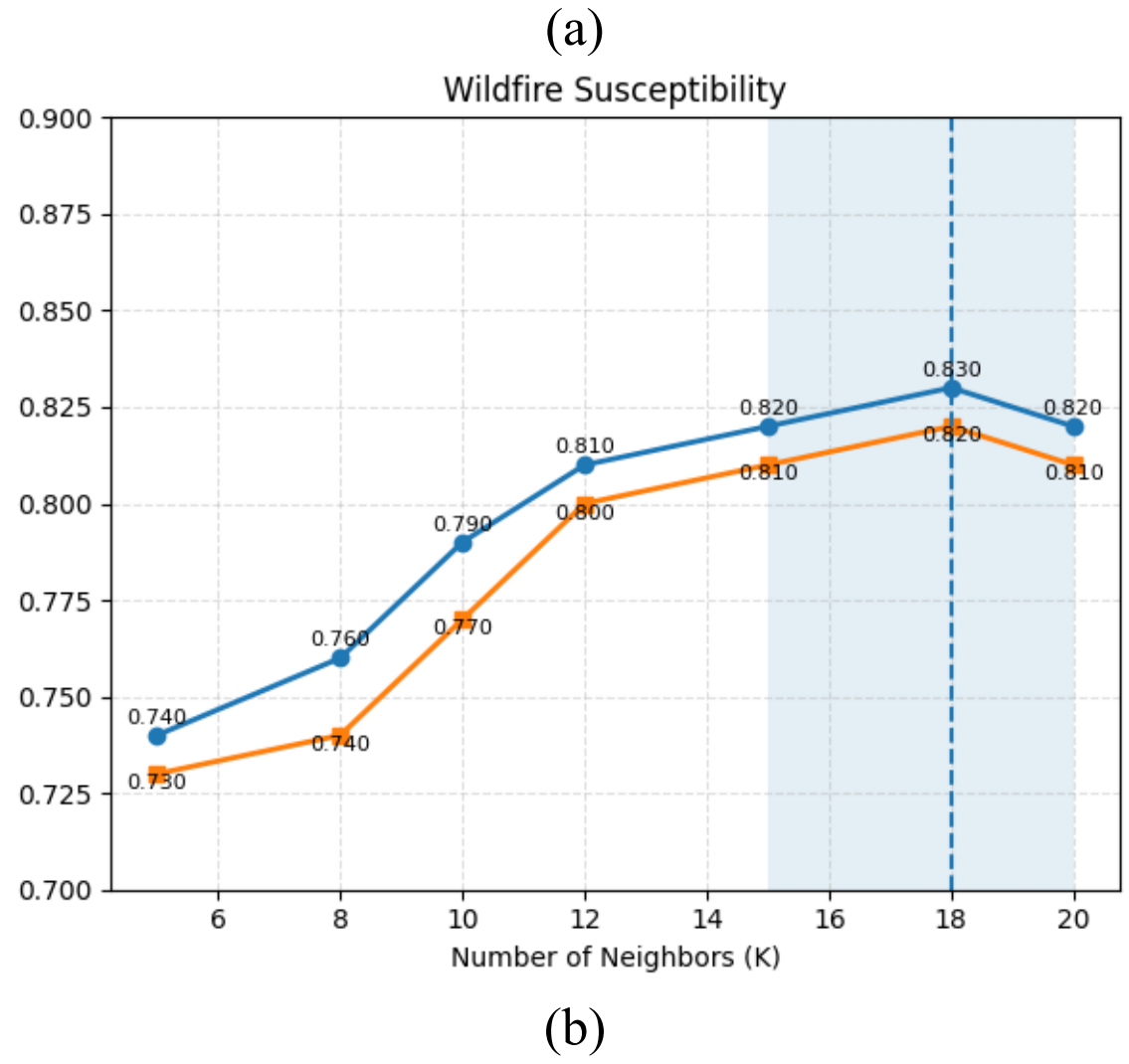

(b)

Figure 12. Sensitivity analysis of the SAGE-XGBoost performance based on the neighborhood value (K): (a) the case of landslide susceptibility; (b) the case of wildfire susceptibility

Table 8. Evaluation of the graph embeddings feature impact in the case of landslide susceptibility

| Method | Accuracy | Precision (Avg.) | Recall (Avg.) | F1-Score (Avg.) |
|---|---|---|---|---|
| **S-XGBoost with Data Augmentation** | 0.83 | 0.83 | 0.82 | 0.81 |
| **SAGE-XGBoost** | 0.88 | 0.88 | 0.86 | 0.87 |

Table 9. Evaluation of the graph embeddings feature impact in the case of wildfire susceptibility

| Method | Accuracy | Precision (Avg.) | Recall (Avg.) | F1-Score (Avg.) |
|---|---|---|---|---|
| **S-XGBoost with Data Augmentation** | 0.78 | 0.80 | 0.77 | 0.78 |
| **SAGE-XGBoost** | 0.83 | 0.83 | 0.82 | 0.82 |

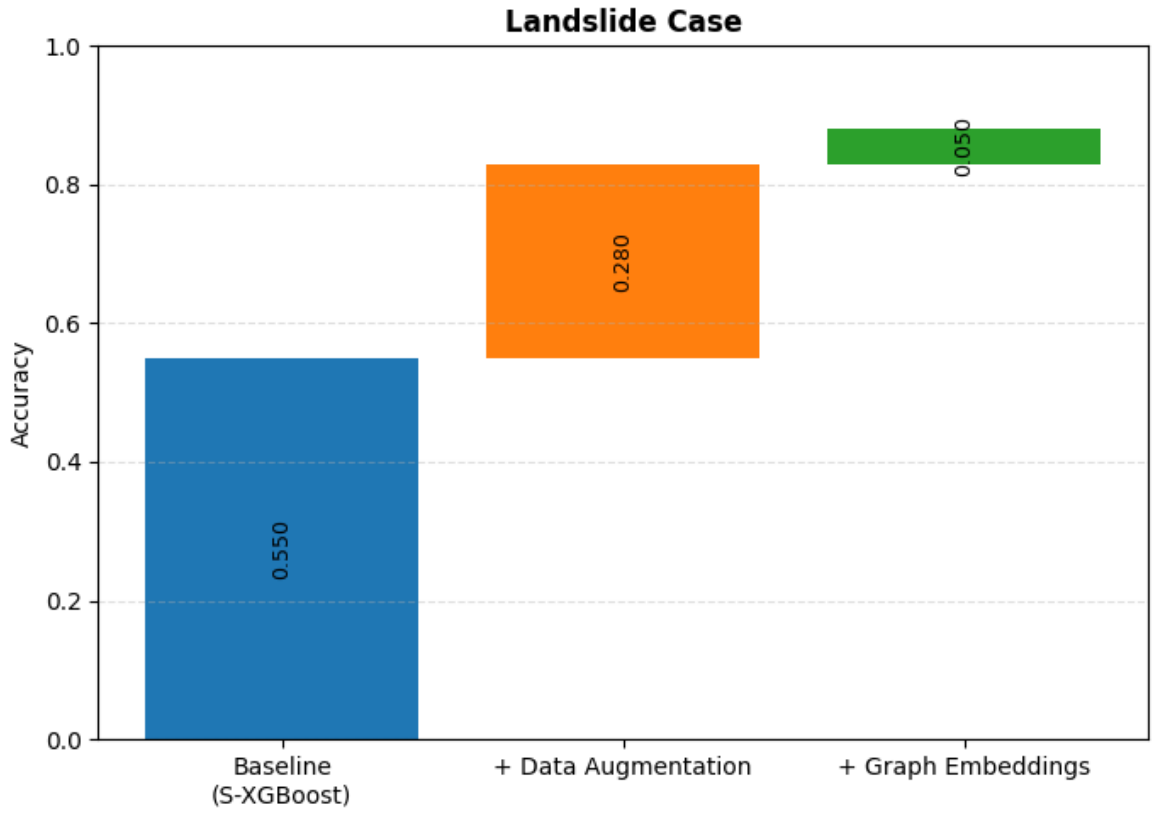

(a)

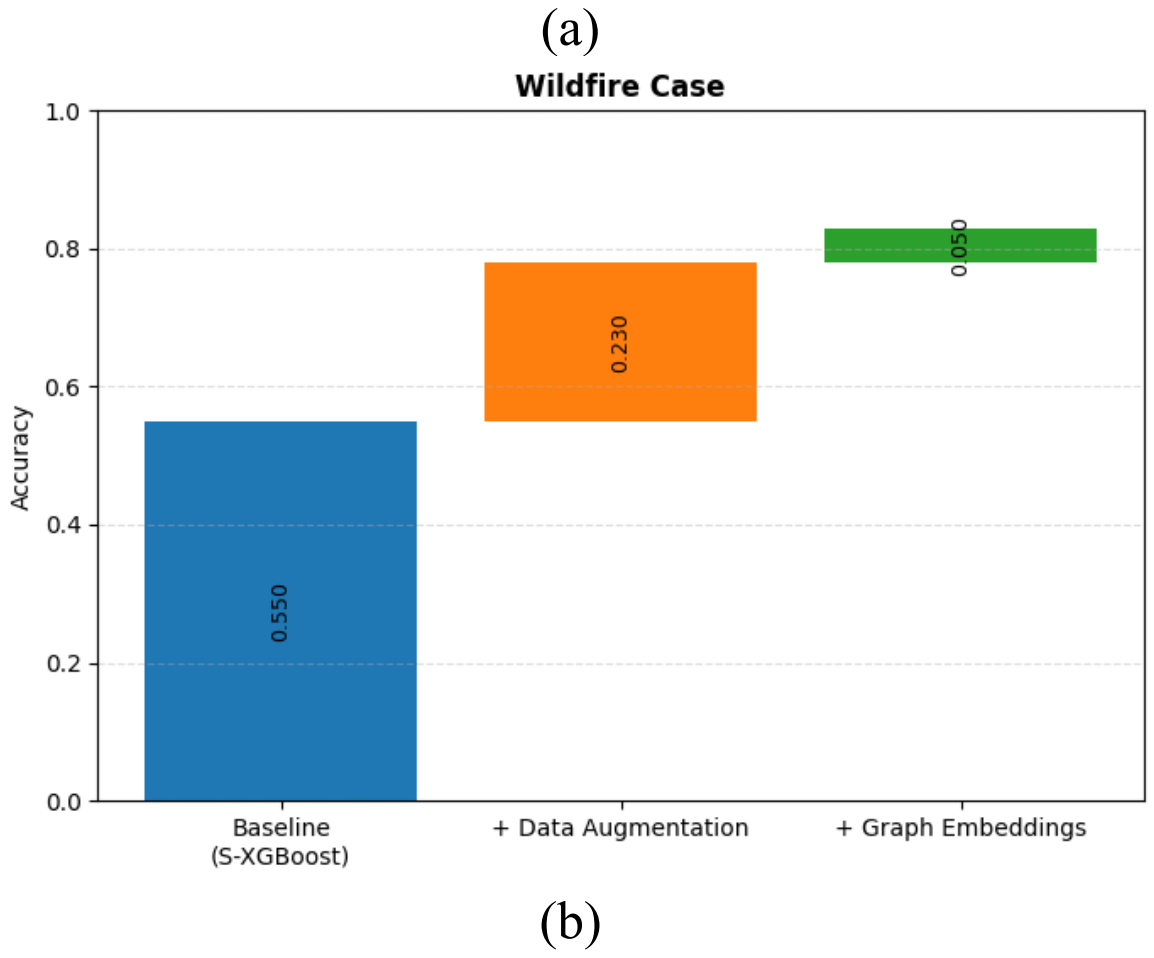

(b)

Figure 13. Waterfall representation of the improvement in performance of the base S-XGboost based on the data augmentation and graph embeddings components: (a) Landslide case; (b) Wildfire case

As illustrated in **Figure 14**, the relative improvement of the ablation variants with respect to the baseline XGBoost model is evident for both landslide and wildfire susceptibility mapping. The findings suggest that each component of the proposed framework contributes positively to model performance, although their impacts are not equal. The incorporation of spatial coordinates yielded only marginal enhancements, while graph embeddings yielded more substantial performance gains in both case studies.

The augmentation of data was found to be the most effective strategy among the individual components, underscoring the significance of addressing the scarcity of data in susceptibility modelling. The full SAGE-XGBoost framework, which integrates spatial coordinates, graph embeddings, and data augmentation, achieved the highest performance in all experiments.

In the context of landslide susceptibility mapping, the complete framework enhanced the accuracy and weighted F1-score by approximately 66% and 71%, respectively, in comparison with the baseline XGBoost model. In a similar vein, for wildfire susceptibility mapping, the corresponding improvements reached approximately 77% for accuracy and 100% for weighted F1-score. These findings indicate that the proposed components are complementary, and that their integration results in a substantially enhanced predictive performance when compared to the performance of any individual component.

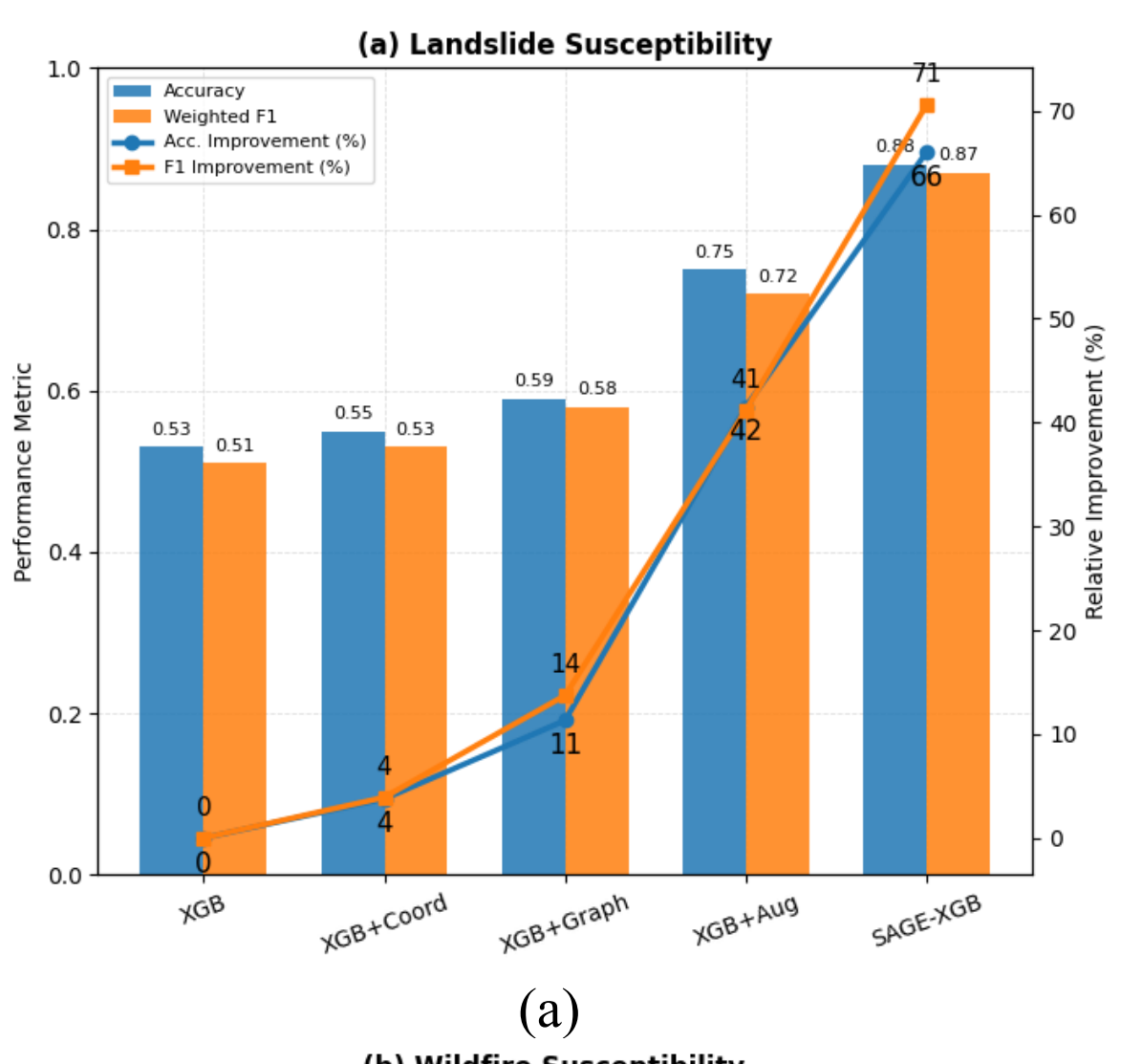

(a)

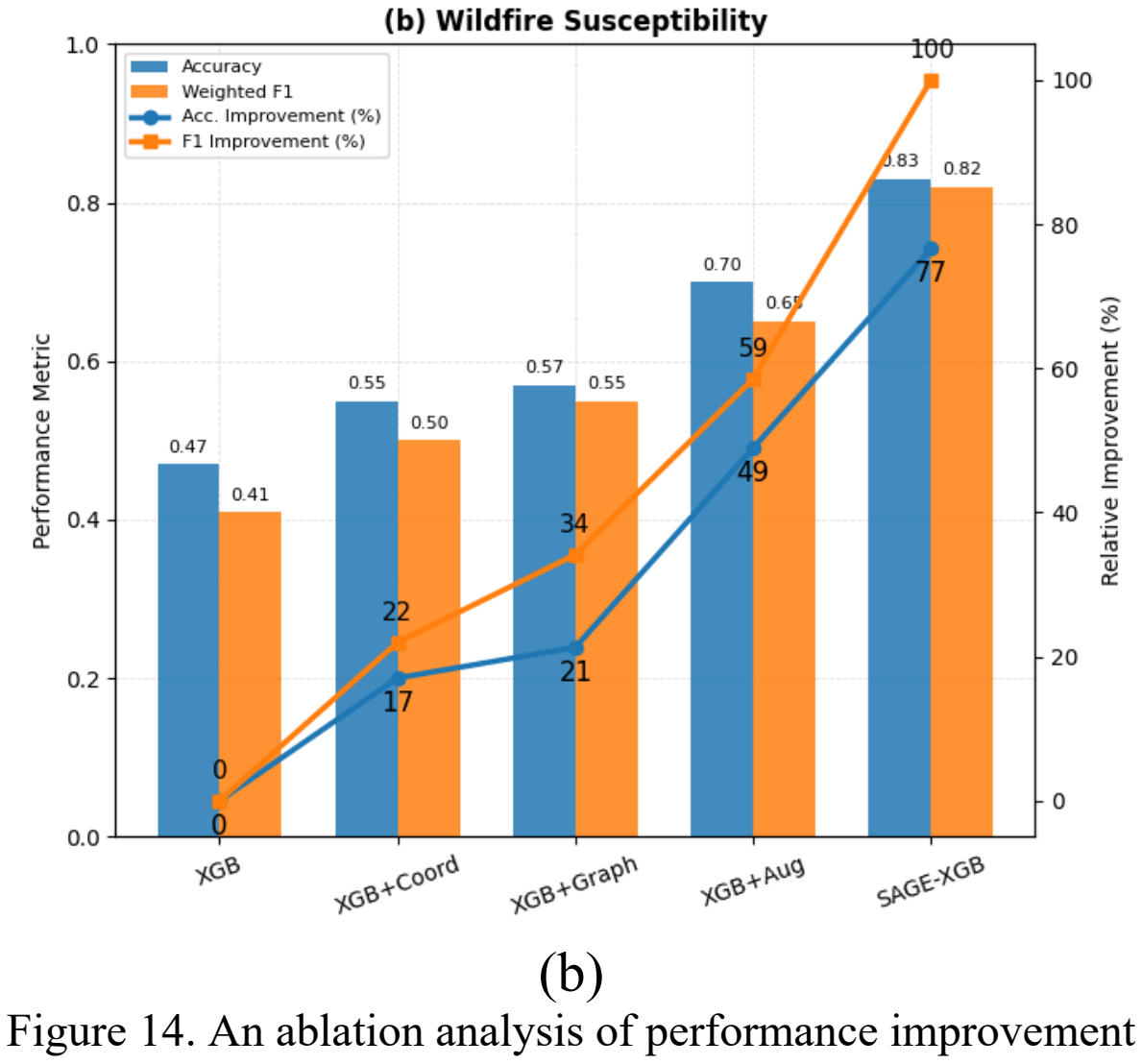

(b)

Figure 14. An ablation analysis of performance improvement based on different components added to the baseline XGBoost: (a) Landslide case; (b) Wildfire case

The gain-based feature importance metric, which measures each feature's contribution to the XGBoost framework's objective (loss) function reduction, is the subject of an additional analysis in this section. The results of this analysis are presented in **Figure 15** for the landslide susceptibility study.

In addition, **Figure 16** indicates the results for the wildfire susceptibility study. In the S-XGBoost model, the feature set encompasses the primary environmental variables, along with the row and column indices that serve as representations of spatial coordinates. In contrast, the SAGE-XGBoost model incorporates these features in addition to three graph embedding components, namely Graph_PC1, Graph_PC2, and Graph_PC3, derived from neighborhood-based structural analysis and dimensionality reduction.

In the landslide case, the gain-based importance results for S-XGBoost identify climate and lithology as two of the most influential environmental predictors. It is noteworthy that in the SAGE-XGBoost model, both lithology and climate remain among the most significant variables, thereby confirming their dominant environmental control on landslide occurrence. However, Graph_PC1 emerges as a highly influential feature, ranking immediately after lithology.

This finding indicates that the graph-based neighborhood analysis, followed by principal component reduction, successfully captures latent local spatial structures that are not explicitly represented by the original environmental variables. In summary, graph embeddings are capable of encoding latent contextual interactions and spatial heterogeneity effects, which contribute significantly to landslide susceptibility prediction.

A similar pattern is observed in the wildfire case. Excluding the spatial coordinate features for the sake of clarity, S-XGBoost identifies humidity and elevation as the most influential environmental predictors. Conversely, within the SAGE-XGBoost framework, elevation and temperature emerge as the two predominant environmental variables. It is noteworthy that Graph_PC2 is identified as a highly influential feature in SAGE-XGBoost. This finding lends further support to the hypothesis that neighborhood-derived graph embeddings are capable of capturing meaningful local spatial interactions and contextual dependencies. These interactions and dependencies appear to enhance model discrimination beyond what can be achieved using environmental covariates alone.

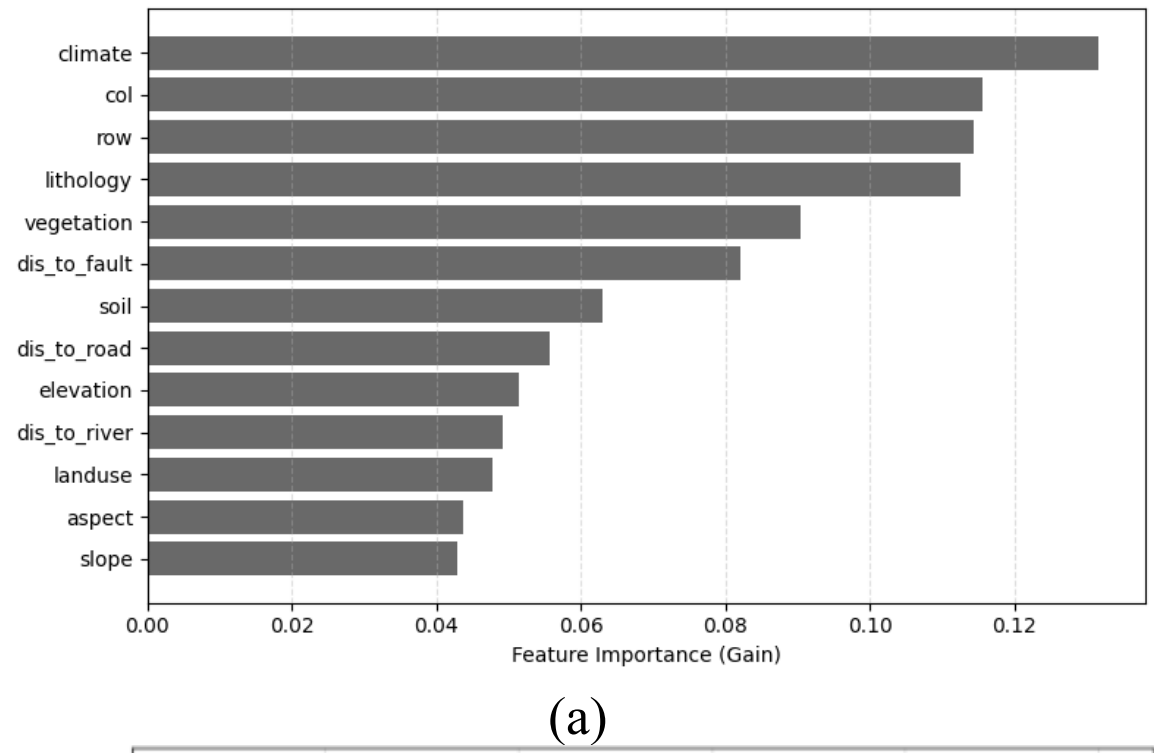

(a)

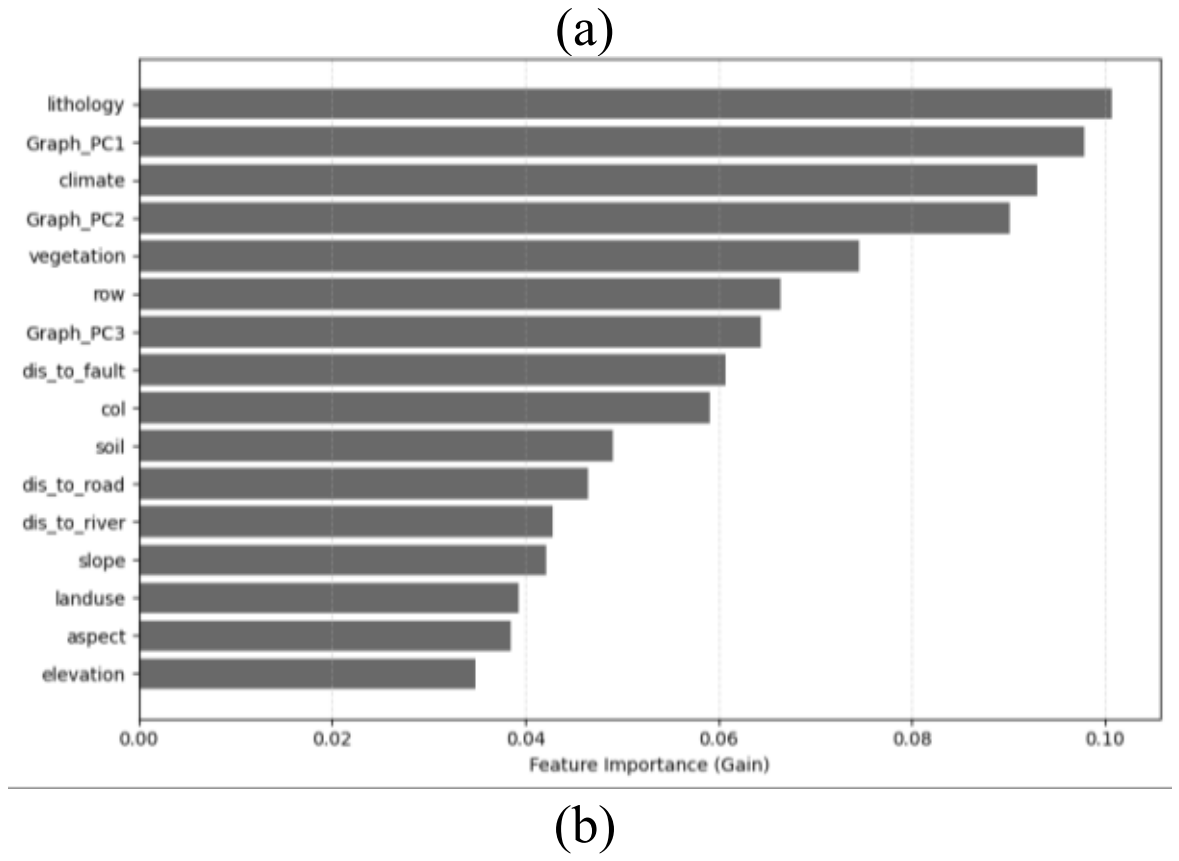

(b)

Figure 15. Feature importance analysis in the landslide susceptibility case study: (a) S-XGBoost; (b): SAGE-XGBoost

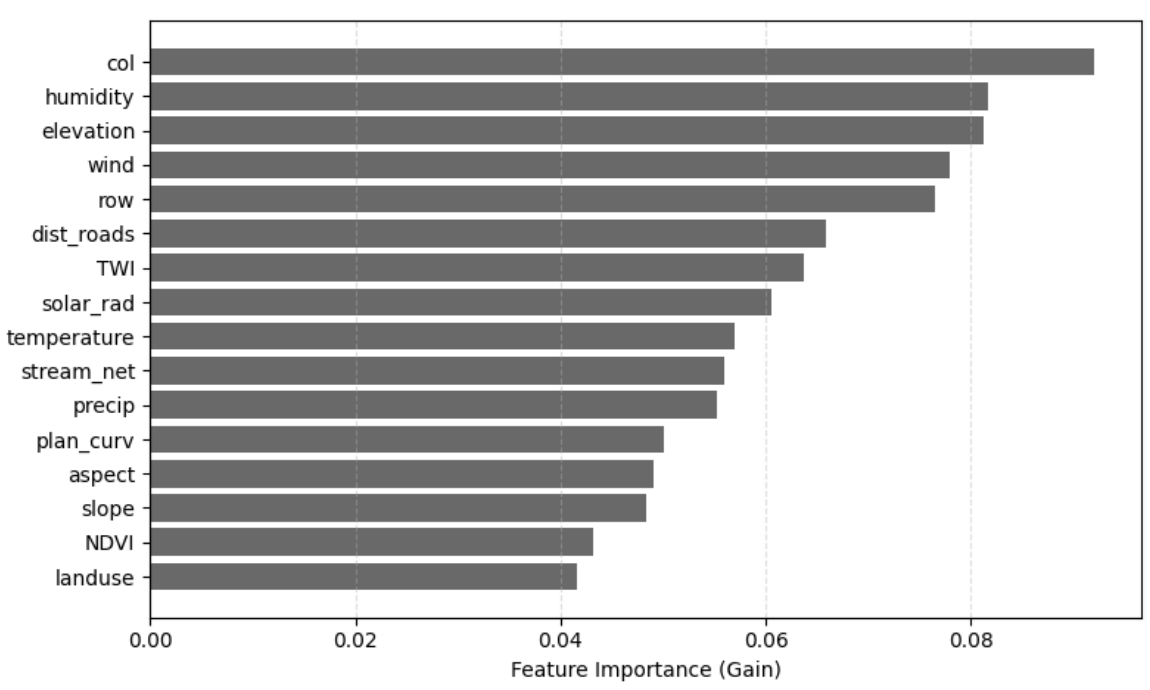

(a)

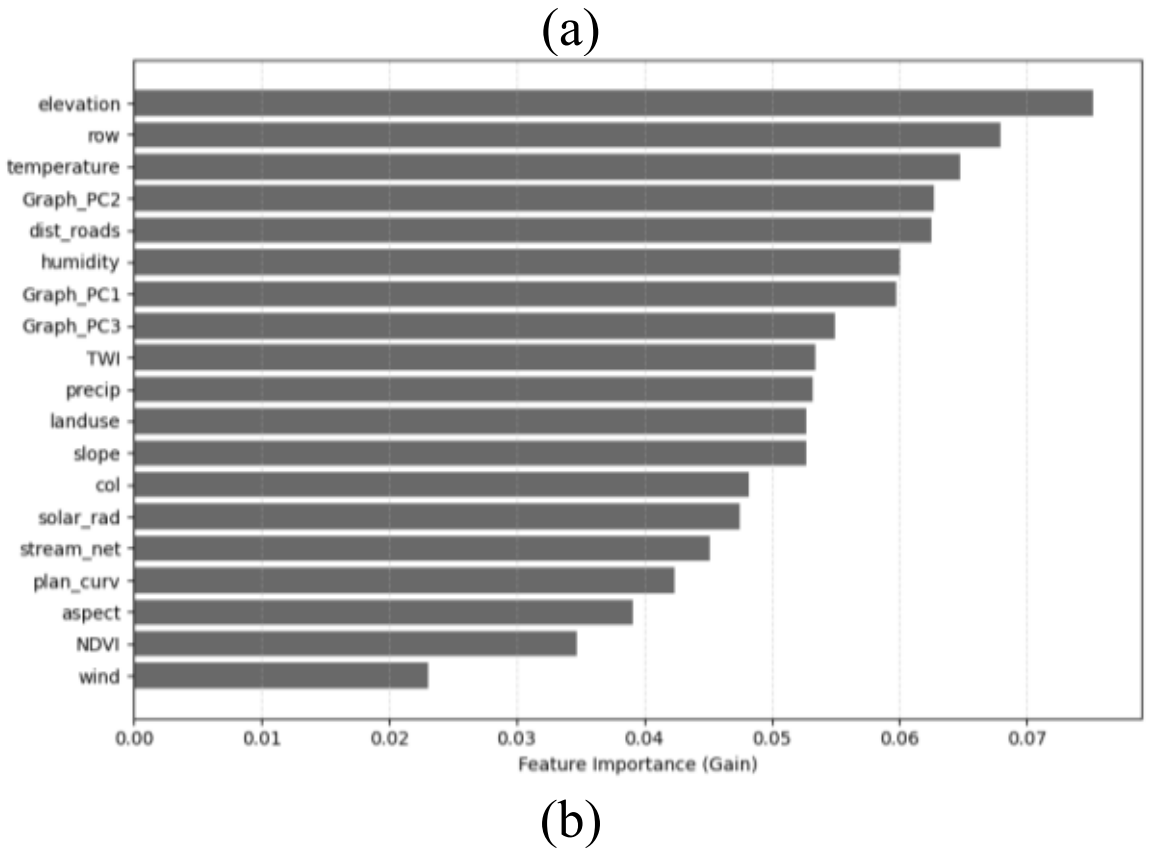

(b)

Figure 16. Feature importance analysis in the wildfire susceptibility case study: (a) S-XGBoost; (b): SAGE-XGBoost

## 5.5. Generalization and Output Susceptibility Maps

Using the best-performing machine learning configurations—namely the enhanced S-XGBoost (improved through the incorporation of the optimal data augmentation strategy) and the proposed SAGE-XGBoost framework, which achieved the highest quantitative performance across all evaluation metrics—continuous susceptibility maps were generated for both case studies. The resulting maps are presented in **Figure 17** for the landslide case and **Figure 18** for the wildfire case.

At a macro scale, both models exhibit analogous global spatial patterns, suggesting that they are equally effective in identifying the most hazardous zones. In the landslide susceptibility assessment in Ardabil Province, areas of high and very high susceptibility are primarily concentrated in the southern region of the study area. In the wildfire case study in the Arasbaran region, elevated susceptibility levels are predominantly observed in central and southern zones.

However, beyond the observed global similarity, significant structural differences emerge in the spatial characteristics of the maps. The outputs of the SAGE-XGBoost model demonstrate a marked enhancement in spatial coherence and a reduction in local discontinuities when compared to the S-XGBoost model. Specifically, transitions between neighboring pixels appear smoother, and abrupt susceptibility fluctuations are substantially mitigated. For the predicted susceptibility maps, Moran's I was computed to offer a numerical evaluation of spatial coherence.

According to the findings, in both case studies, SAGE-XGBoost generated slightly highermore spatial autocorrelation than Spatial-XGBoost. Moran's I for mapping landslide susceptibility rose from 0.70 for Spatial-XGBoost to 0.72 for SAGE-XGBoost. In a similar vein, Moran's I rose from 0.68 to 0.74 for wildfire susceptibility mapping. These findings imply that while maintaining the general spatial structure of the predictions, the graph-based embedding approach helps to produce more spatially continuous susceptibility patterns with less local fragmentation.

From a spatial modeling perspective, this enhancement can be ascribed to the incorporation of graph embedding features, which implicitly encode neighborhood-based contextual information. The incorporation of graph-derived principal components within the model facilitates the capture of latent spatial autocorrelation structures and higher-order local interactions that are not explicitly represented in the original environmental covariates. This mechanism functions as a form of data-driven spatial regularization, thereby reducing noise amplification and limiting over-sensitive local variations that may arise in purely feature-based models. The impact is especially pronounced in the central and southern regions of both case studies, where SAGE-XGBoost generates more spatially consistent and geologically plausible susceptibility gradients. This finding indicates an augmented spatial generalization capability and enhanced robustness under data-scarce conditions.

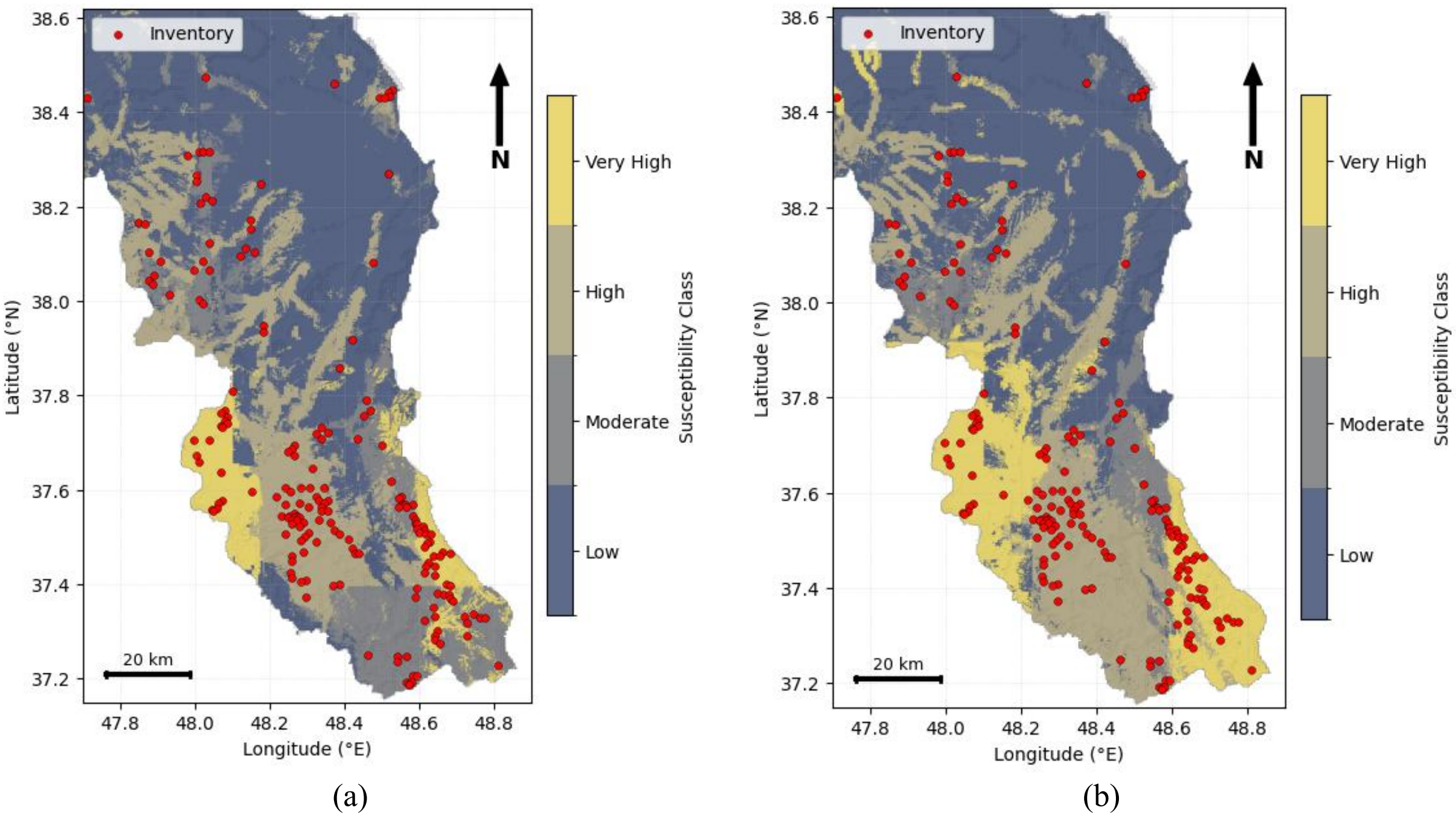

Figure 17. Landslide susceptibility maps from the selected machine learning methods: (a) S-XGBoost with data augmentation; (b) SAGE-XGBoost

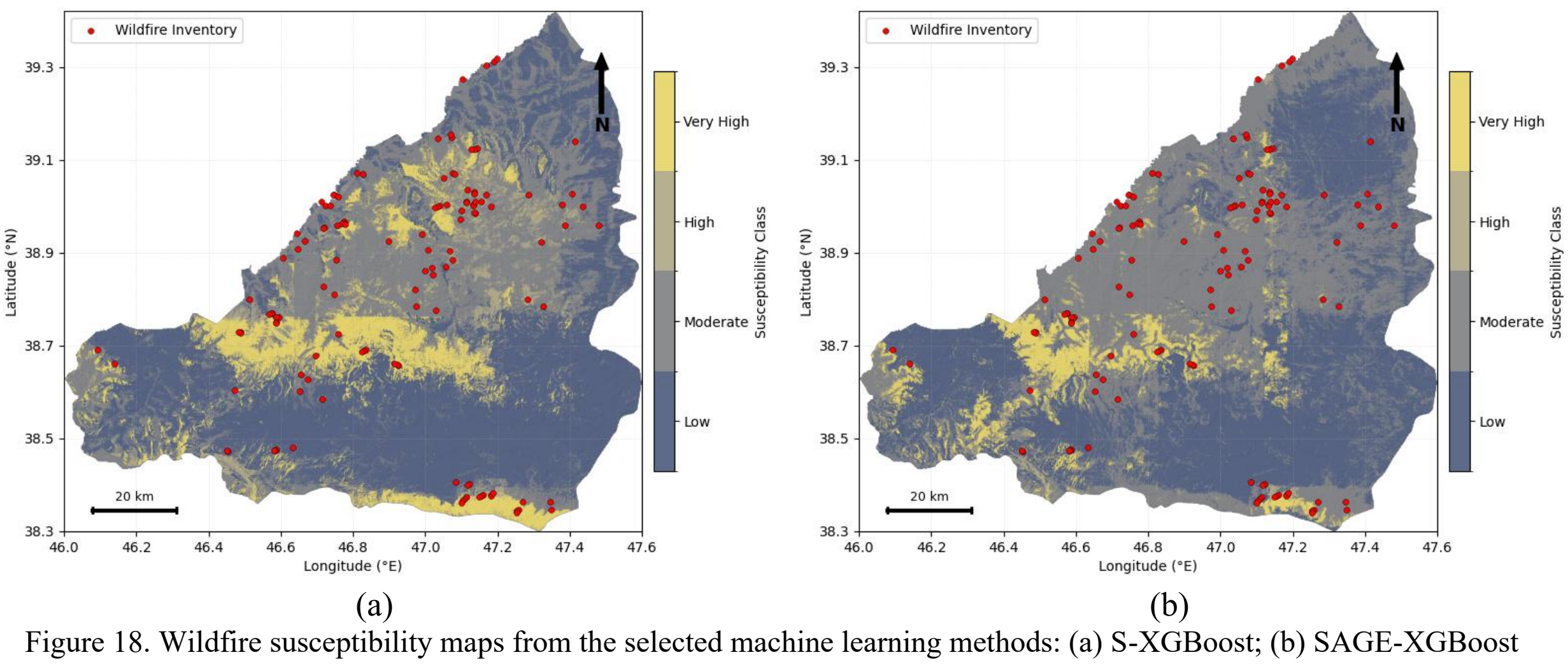

Figure 18. Wildfire susceptibility maps from the selected machine learning methods: (a) S-XGBoost; (b) SAGE-XGBoost

## 5.6. Computational Efficiency

To determine whether the suggested methodology could be used for large-scale susceptibility mapping in practice, its computational efficiency was also assessed. For the landslide study area with 10,598,347 valid 30-m raster pixels, training time, peak memory consumption, and full-map prediction time were compared.

**Table 10** demonstrates that XGBoost had the shortest training time (31.03 s), while the suggested SAGE-XGBoost needed 57.23 s because of the extra graph-feature creation and embedding steps. Despite this increase, the training time was still comparable to that of a Random Forest (60.68 s). All approaches had low memory utilization, with the experimental environment's peak memory usage being below 1 MB.

Prediction time for full-area susceptibility mapping increased steadily as model complexity grew. XGBoost produced the susceptibility map in 183.96 s, Spatial-XGBoost in 363.98 s, and SAGE-XGBoost in 1,262.40 s (approximately 21 minutes), including graph-feature synthesis, PCA embedding, and model inference. For offline susceptibility mapping applications where model quality is more important than real-time prediction speed, this increased computational cost might be justified given the significant boost in predictive performance attained by SAGE-XGBoost.

Table 10. Comparison of computational efficiency and memory usage of different methods

| | Random Forest | XGBoost | Spatial-XGBoost | SAGE-XGBoost |
|---|---|---|---|---|
| **Training Time (s)** | 60.68 | 31.03 | 32.84 | 57.23 |
| **Peak Memory Usage (MB)** | 0.85 | 0.20 | 0.24 | 0.28 |
| **Prediction Time (By accounting Feature Generation) (s)** | 155.85 | 183.96 | 363.98 | 1262.40 |

## 5.7. Comparison, Limitations, and Future Directions

One widely accepted criterion for comparing susceptibility modeling studies is the AUC, a threshold-insensitive measure of performance, frequently adopted as a benchmark metric in environmental hazard assessment. Although the baseline models in this study produced relatively modest AUC values, remaining below 0.77 in the best-performing non-spatial approaches, the proposed SAGE-XGBoost framework demonstrated substantial improvement. In the landslide case study, SAGE-XGBoost achieved an AUC of approximately 0.97, indicating excellent discriminatory capability.

By comparison, Ado et al. (2022) applied multiple machine learning and deep learning techniques and reported a maximum AUC of 0.90. Kadavi et al. (2018) employed hybrid ensemble techniques, including LogitBoost, AdaBoost, Bagging, and Multiclass Classifier models, and achieved an optimal AUC of 0.86. Azarafza et al. (2021) used a deep convolutional neural network to assess landslide susceptibility and reported a maximum AUC of 0.91. Xiao et al. (2025) compared logistic regression (LR), random forest (RF), and graph neural networks (GNN). The GNN method demonstrated the highest performance, attaining an AUC of 0.87. Only a few studies, such as that of Agboola et al. (2024), reported AUC values of 0.91 and 0.99 under different optimization scenarios. However, these results were obtained using a highly enriched dataset of 1,215 historical landslide occurrences and 1,215 non-landslide samples. This indicates substantially lower data scarcity conditions than in the present study. It should be noted that direct cross-study comparisons of AUC values should be interpreted with caution, as variations in characteristics of the study area, spatial resolution, inventory quality, sample size, and validation strategies can significantly impact predictive performance.

In the context of wildfire susceptibility mapping, Bjånes et al. (2021) reported the highest AUC of 0.95 using an ensemble deep learning approach. Subsequent to this, convolutional neural network (CNN)-1 was implemented, yielding an AUC of 0.90. Gholamnia et al. (2020) evaluated ten machine learning algorithms alongside logistic regression and achieved an optimal AUC of 0.88 with Random Forest. Bahadori et al. (2023) incorporated temporal dynamics using deep learning architectures, such as LSTM and RNN. Based on a dataset of about 600 wildfire locations derived from Landsat-8 imagery and 232 locations from MODIS imagery, they achieved an optimal AUC of 0.97. Durlević et al. (2025) similarly reported an optimal AUC of 0.92 using deep neural networks and XGBoost.

These comparisons highlight the competitive performance of SAGE-XGBoost despite operating under explicit data scarcity conditions and without the use of deep learning methods. The proposed framework was successfully validated for both landslide and wildfire susceptibility prediction, demonstrating that structurally informed feature engineering and controlled augmentation can compensate for limited labeled data. Beyond these two hazards, the framework has strong potential to generalize to other environmental risks, such as floods, debris flows, and droughts. More broadly, the methodology has the potential to offer a transferable approach to modelling spatial heterogeneity in various geospatial prediction problems.

While the proposed framework is predominantly data-driven, the incorporation of geomechanical, hydrological, or process-based knowledge into the feature construction and learning stages may prove advantageous in future developments (Taherdangkoo et al., 2026; Barella et al., 2026; Cui et al., 2026). In circumstances where data is scarce, where purely data-driven models may encounter limitations due to an insufficient number of observations, the utilization of hybrid approaches holds the potential to enhance physical consistency, uncertainty characterization, and prediction reliability.

This study modeled spatial relationships for graph embeddings using a K-nearest neighbor (KNN) approach and implemented data augmentation through a controlled, noise-based strategy. Unlike deep representation learning approaches, which require large labeled datasets to learn latent spatial representations end-to-end, the proposed SAGE framework uses a structural learning paradigm that explicitly encodes neighborhood-based spatial structures into compact graph embeddings. The spatial context incorporated within the graph-based features is governed by the neighborhood size K. While larger K values may diminish the strict locality of neighborhood relationships, sensitivity analysis has demonstrated that model performance remains relatively stable within a moderate range of K values. Under data-scarce circumstances, the optimal values (K = 15 for landslide susceptibility and K = 18 for wildfire susceptibility) strike a balance between capturing local environmental similarities and preserving sufficient neighborhood information. To further enhance locality preservation, future research may consider investigating alternative spatial graph construction techniques, and adaptive or density-dependent neighborhood definitions (Vahidnia, 2025).

Furthermore, this study did not use the variables known as lithology, soil type, land use, and related criteria as nominal category identifiers. Rather, these characteristics were converted into ranked classes according to their respective impact on the target hazard phenomenon within the study area prior to model building. As a result, rather than arbitrary categorical labels, the encoded values represent an ordinal susceptibility ranking. Future research may use a dual augmentation method, wherein category-specific augmentation techniques are used to handle categorical variables and noise-based approaches are used to modify continuous data, along with advanced augmentation techniques designed specifically for tabular data (Machado et al., 2022).

It is imperative to acknowledge that the term data scarcity in this study signifies the paucity of labeled target samples. Additional challenges related to data are represented by missing data, particularly in time-dependent datasets (Emmanuel et al., 2021)—represent additional constraints that warrant further investigation. Future research endeavors may extend the SAGE-XGBoost framework to address broader dimensions of data limitation, thereby enabling a more comprehensive treatment of data scarcity issues in environmental susceptibility modeling.

## 6. Conclusion

In this study, a novel machine learning framework (SAGE-XGBoost) was proposed for natural hazards susceptibility mapping under data-scarce conditions. Two case studies—landslide and wildfire susceptibility—were employed to test the underlying hypotheses. In addition to conventional machine learning approaches, spatially explicit methods capable of modeling spatial heterogeneity were considered for comparison. These methods include Geographically Weighted Random Forest (GWRF) and Spatial XGBoost (S-XGBoost). The primary conclusions that can be drawn from these findings are as follows:

SAGE-XGBoost demonstrated consistent superiority over all competing methods across all evaluation metrics. In both case studies, the superiority of the proposed method over S-XGBoost exceeded 33%, while the improvement in performance compared to non-spatial baseline machine learning models was approximately 40%, as measured by performance indicators.

According to the sensitivity analysis, optimal results were achieved when the noise-based data augmentation rate ranged between 1.3 and 1.5 of the original dataset size. This controlled augmentation strategy not only enhanced predictive performance but also prevented overfitting, avoided unnecessary expansion of synthetic samples, and minimized additional computational burden during model training. Moreover, by condensing latent neighborhood information into a small number of principal components (three features based on the sensitivity analysis), the SAGE framework obviates the necessity for computationally intensive deep learning architectures while maintaining robust representational capacity.

Both core components of the proposed framework—noise-based data augmentation and graph embedding-based feature generation—play constructive roles in improving model performance. The gain-based feature importance analysis substantiates that, while pivotal environmental variables maintain their foundational influence in both case studies, graph embedding components in SAGE furnish augmented structurally informed explanatory power. The graph embeddings, derived from neighborhood statistics including mean, standard deviation, distance-weighted mean, and local density (with a maximum neighborhood size of 20 points), contributed at least a 5% improvement in predictive performance. This finding lends further support to the hypothesis that integrating neighborhood-based contextual information enables the model to capture latent spatial mechanisms underlying environmental susceptibility processes.

The integration of graph embeddings not only improves classification metrics but also enhances the structural realism, continuity, and spatial stability of the generated susceptibility maps. By reducing noise amplification and capturing latent spatial autocorrelation patterns, the proposed SAGE-XGBoost framework demonstrates enhanced spatial generalization and practical reliability for environmental hazard assessment in circumstances where target data is limited.

## References

Abedi Gheshlaghi, H., Feizizadeh, B., Blaschke, T., Lakes, T., & Tajbar, S. (2021). Forest fire susceptibility modeling using hybrid approaches. *Transactions in GIS*, *25*(1), 311-333.

Abedi, R., Costache, R., Shafizadeh-Moghadam, H., & Pham, Q. B. (2022). Flash-flood susceptibility mapping based on XGBoost, random forest and boosted regression trees. *Geocarto International*, *37*(19), 5479-5496.

Ado, M., Amitab, K., Maji, A. K., Jasińska, E., Gono, R., Leonowicz, Z., & Jasiński, M. (2022). Landslide susceptibility mapping using machine learning: A literature survey. *Remote Sensing*, *14*(13), 3029.

Agboola, G., Beni, L. H., Elbayoumi, T., & Thompson, G. (2024). Optimizing landslide susceptibility mapping using machine learning and geospatial techniques. *Ecological Informatics*, *81*, 102583.

Anil, D., & Manjula, S. H. (2025). High-Precision Landslide Susceptibility Mapping Using CNN-LSTM-Attention Models. *Engineering, Technology & Applied Science Research*, *15*(4), 25486-25491.

Azarafza, M., Azarafza, M., Akgün, H., Atkinson, P. M., & Derakhshani, R. (2021). Deep learning-based landslide susceptibility mapping. *Scientific reports*, *11*(1), 24112.

Ba, Q., Chen, Y., Deng, S., Wu, Q., Yang, J., & Zhang, J. (2017). An improved information value model based on gray clustering for landslide susceptibility mapping. *ISPRS International Journal of Geo-Information*, *6*(1), 18.

Bahadori, N., Razavi-Termeh, S. V., Sadeghi-Niaraki, A., Al-Kindi, K. M., Abuhmed, T., Nazeri, B., & Choi, S. M. (2023). Wildfire susceptibility mapping using deep learning algorithms in two satellite imagery dataset. *Forests*, *14*(7), 1325.

Barella, C. F., Zêzere, J. L., & Fernandes, N. F. (2026). Validation and prediction challenges in data-driven landslide susceptibility mapping: insights from random sampling of training and test data in machine learning models. *Bulletin of Engineering Geology and the Environment*, *85*(4), 253.

Belussi, A., Migliorini, S., & Eldawy, A. (2024). A Generic Machine Learning Model for Spatial Query Optimization based on Spatial Embeddings. *ACM Transactions on Spatial Algorithms and Systems*, *10*(4), 1-33.

Bjånes, A., De La Fuente, R., & Mena, P. (2021). A deep learning ensemble model for wildfire susceptibility mapping. *Ecological Informatics*, *65*, 101397.

Can, R., Kocaman, S., & Gokceoglu, C. (2021). A comprehensive assessment of XGBoost algorithm for landslide susceptibility mapping in the upper basin of Ataturk dam, Turkey. *Applied Sciences*, *11*(11), 4993.

Chen, W., Li, W., Chai, H., Hou, E., Li, X., & Ding, X. (2016). GIS-based landslide susceptibility mapping using analytical hierarchy process (AHP) and certainty factor (CF) models for the Baozhong region of Baoji City, China. *Environmental Earth Sciences*, *75*(1), 63.

Chihi, H., Hammami, M. A., & Mezni, I. (2025). Flood susceptibility mapping in data-scarce arid environments: guided by geology-driven knowledge and multi-event cloud-based validation. *Natural Hazards*, *121*(18), 20855-20901.

Cui, H., Pei, T., Devineni, N., Tian, Y., Shen, C., & Ji, J. (2026). A multi-objective physics-informed machine learning framework for landslide susceptibility mapping. *Georisk: Assessment and Management of Risk for Engineered Systems and Geohazards*, 1-25.

Dano, U. L., Balogun, A. L., Matori, A. N., Wan Yusouf, K., Abubakar, I. R., Said Mohamed, M. A., ... & Pradhan, B. (2019). Flood susceptibility mapping using GIS-based analytic network process: A case study of Perlis, Malaysia. *Water*, *11*(3), 615.

Du, J., Wang, S., Ye, X., Sinton, D. S., & Kemp, K. (2022). GIS-KG: Building a large-scale hierarchical knowledge graph for geographic information science. *International Journal of Geographical Information Science*, *36*(5), 873-897.

Durlević, U., Ilić, V., & Valjarević, A. (2025). Wildfire susceptibility mapping using deep learning and machine learning models based on multi-sensor satellite data fusion: a case study of Serbia. *Fire*, *8*(10), 407.

El Miloudi, Y., El Kharim, Y., & El Hamdouni, R. (2025). A novel approach for rockfall susceptibility mapping: Transfer learning between boosting models and logistic regression. *Environmental Earth Sciences*, *84*(16), 447.

Emmanuel, T., Maupong, T., Mpoeleng, D., Semong, T., Mphago, B., & Tabona, O. (2021). A survey on missing data in machine learning. *Journal of Big data*, *8*(1), 140.

Falah, F., Rahmati, O., Rostami, M., Ahmadisharaf, E., Daliakopoulos, I. N., & Pourghasemi, H. R. (2019). Artificial neural networks for flood susceptibility mapping in data-scarce urban areas. In *Spatial modeling in GIS and R for Earth and Environmental Sciences* (pp. 323-336). Elsevier.

Gao, R., Wang, C., Wu, D., Liu, H., & Liu, X. (2025). Comprehensive application of transfer learning, unsupervised learning and supervised learning in debris flow susceptibility mapping. *Applied Soft Computing*, *170*, 112612.

Gholamnia, K., Gudiyangada Nachappa, T., Ghorbanzadeh, O., & Blaschke, T. (2020). Comparisons of diverse machine learning approaches for wildfire susceptibility mapping. *Symmetry*, *12*(4), 604.

Gigović, L., Pourghasemi, H. R., Drobnjak, S., & Bai, S. (2019). Testing a new ensemble model based on SVM and random forest in forest fire susceptibility assessment and its mapping in Serbia's Tara National Park. *Forests*, *10*(5), 408.

Ha, H., Bui, Q. D., Nguyen, H. D., Pham, B. T., Lai, T. D., & Luu, C. (2023). A practical approach to flood hazard, vulnerability, and risk assessing and mapping for Quang Binh province, Vietnam. *Environment, Development and Sustainability*, *25*(2), 1101-1130.

Hamilton, W., Ying, Z., & Leskovec, J. (2017). Inductive representation learning on large graphs. *Advances in neural information processing systems*, *30*.

Ilia, I., & Tsangaratos, P. (2016). Applying weight of evidence method and sensitivity analysis to produce a landslide susceptibility map. *Landslides*, *13*(2), 379-397.

Jahanbani, M., Vahidnia, M. H., Aghamohammadi, H., & Azizi, Z. (2024). Flood susceptibility mapping through geoinformatics and ensemble learning methods, with an emphasis on the AdaBoost-Decision Tree algorithm, in Mazandaran, Iran. *Earth Science Informatics*, *17*(2), 1433-1457.

Kadavi, P. R., Lee, C. W., & Lee, S. (2018). Application of ensemble-based machine learning models to landslide susceptibility mapping. *Remote Sensing*, *10*(8), 1252.

Kanani-Sadat, Y., Arabsheibani, R., Karimipour, F., & Nasseri, M. (2019). A new approach to flood susceptibility assessment in data-scarce and ungauged regions based on GIS-based hybrid multi criteria decision-making method. *Journal of hydrology*, *572*, 17-31.

Kong, L., Feng, W., Yi, X., Xue, Z., & Bai, L. (2025). Enhanced landslide susceptibility mapping in data-scarce regions via unsupervised few-shot learning. *Gondwana Research*, *138*, 31-46.

Kumar, S., Snehmani, Srivastava, P. K., Gore, A., & Singh, M. K. (2016). Fuzzy–frequency ratio model for avalanche susceptibility mapping. *International Journal of Digital Earth*, *9*(12), 1168-1184.

Lee, S. (2004). Application of likelihood ratio and logistic regression models to landslide susceptibility mapping using GIS. *Environmental Management*, *34*(2), 223-232.

Li, Y., Jiang, W., Feng, X., Lv, S., Yu, W., & Ma, E. (2024). Debris flow susceptibility mapping in alpine canyon region: a case study of Nujiang Prefecture. *Bulletin of Engineering Geology and the Environment*, *83*(5), 169.

Lu, F., Zhang, G., Wang, T., Ye, Y., & Zhao, Q. (2025). Geographically Weighted Random Forest Based on Spatial Factor Optimization for the Assessment of Landslide Susceptibility. *Remote Sensing*, *17*(9), 1608.

Machado, P., Fernandes, B., & Novais, P. (2022, November). Benchmarking data augmentation techniques for tabular data. In *International Conference on Intelligent Data Engineering and Automated Learning* (pp. 104-112). Cham: Springer International Publishing.

Mosavi, A., Shirzadi, A., Choubin, B., Taromideh, F., Hosseini, F. S., Borji, M., ... & Dineva, A. A. (2020). Towards an ensemble machine learning model of random subspace based functional tree classifier for snow avalanche susceptibility mapping. *IEEE Access*, *8*, 145968-145983.

Nachappa, T. G., Piralilou, S. T., Gholamnia, K., Ghorbanzadeh, O., Rahmati, O., & Blaschke, T. (2020). Flood susceptibility mapping with machine learning, multi-criteria decision analysis and ensemble using Dempster Shafer Theory. *Journal of hydrology*, *590*, 125275.

Ngo, P. T. T., Panahi, M., Khosravi, K., Ghorbanzadeh, O., Kariminejad, N., Cerda, A., & Lee, S. (2021). Evaluation of deep learning algorithms for national scale landslide susceptibility mapping of Iran. *Geoscience Frontiers*, *12*(2), 505-519.

Pourghasemi, H. R., Teimoori Yansari, Z., Panagos, P., & Pradhan, B. (2018). Analysis and evaluation of landslide susceptibility: a review on articles published during 2005–2016 (periods of 2005–2012 and 2013–2016). *Arabian Journal of Geosciences*, *11*(9), 193.

Pradhan, B., Lee, S., Dikshit, A., & Kim, H. (2023). Spatial flood susceptibility mapping using an explainable artificial intelligence (XAI) model. *Geoscience Frontiers*, *14*(6), 101625.

Pugliese Viloria, A. D. J., Folini, A., Carrion, D., & Brovelli, M. A. (2024). Hazard susceptibility mapping with machine and deep learning: a literature review. *Remote Sensing*, *16*(18), 3374.

Rahmati, O., Pourghasemi, H. R., & Zeinivand, H. (2016). Flood susceptibility mapping using frequency ratio and weights-of-evidence models in the Golastan Province, Iran. *Geocarto International*, *31*(1), 42-70.

Reichenbach, P., Rossi, M., Malamud, B. D., Mihir, M., & Guzzetti, F. (2018). A review of statistically-based landslide susceptibility models. *Earth-science reviews*, *180*, 60-91.

Rezaie, F., Panahi, M., Jun, C., Dayal, K., Kim, D., Darabi, H., ... & Bateni, S. M. (2025). Deep learning models for drought susceptibility mapping in Southeast Queensland, Australia. *Stochastic Environmental Research and Risk Assessment*, 1-17.

Ribeiro, L., Pindo, J. C., & Dominguez-Granda, L. (2017). Assessment of groundwater vulnerability in the Daule aquifer, Ecuador, using the susceptibility index method. *Science of the total environment*, *574*, 1674-1683.

Roberts, D. R., Bahn, V., Ciuti, S., Boyce, M. S., Elith, J., Guillera-Arroita, G., ... & Dormann, C. F. (2017). Cross-validation strategies for data with temporal, spatial, hierarchical, or phylogenetic structure. *Ecography*, *40*(8), 913-929.

Tacconi Stefanelli, C., Casagli, N., & Catani, F. (2020). Landslide damming hazard susceptibility maps: a new GIS-based procedure for risk management. *Landslides*, *17*(7), 1635-1648.

Taherdangkoo, R., Franz, L. S., Arab, A., Wichert, J., & Butscher, C. (2026). Reliable probabilistic landslide susceptibility mapping using a calibrated stacked ensemble with bootstrap-based uncertainty. *Bulletin of Engineering Geology and the Environment*, *85*(5), 333.

Tehrany, M. S., Pradhan, B., Mansor, S., & Ahmad, N. (2015). Flood susceptibility assessment using GIS-based support vector machine model with different kernel types. *Catena*, *125*, 91-101.

Thiery, Y., Malet, J. P., Sterlacchini, S., Puissant, A., & Maquaire, O. (2007). Landslide susceptibility assessment by bivariate methods at large scales: application to a complex mountainous environment. *Geomorphology*, *92*(1-2), 38-59.

Vahidnia, M. H. (2025). ADT-GWR: adaptive Delaunay topology-based geographically weighted regression. *Spatial Information Research*, *33*(5), 41.

Vahidnia, M. H., Alesheikh, A. A., Alimohammadi, A., & Hosseinali, F. (2010). A GIS-based neuro-fuzzy procedure for integrating knowledge and data in landslide susceptibility mapping. *Computers & Geosciences*, *36*(9), 1101-1114.

Van Westen, C. J., Castellanos, E., & Kuriakose, S. L. (2008). Spatial data for landslide susceptibility, hazard, and vulnerability assessment: An overview. *Engineering geology*, *102*(3-4), 112-131.

Xiao, T., Huang, W., Wang, L., Yang, B., Qin, Z., Liu, X., & Xiao, Y. (2025). Uncertainty-aware ensemble learning and dynamic threshold optimization for landslide susceptibility mapping. *Computers & Geosciences*, 106042.

Yan, Y., Chen, R., Zhang, Y., Zhang, H., Cai, Y., & He, B. (2025, August). Integrating Long Short-Term Memory Networks and Machine Learning Models for Improving Explainable Wildfire Susceptibility Assessment. In *IGARSS 2025-2025 IEEE International Geoscience and Remote Sensing Symposium* (pp. 3126-3130). IEEE.

Youssef, A. M., Pradhan, B., Dikshit, A., Al-Katheri, M. M., Matar, S. S., & Mahdi, A. M. (2022). Landslide susceptibility mapping using CNN-1D and 2D deep learning algorithms: comparison of their performance at Asir Region, KSA. *Bulletin of Engineering Geology and the Environment*, *81*(4), 165.

Zhao, P., Masoumi, Z., Kalantari, M., Aflaki, M., & Mansourian, A. (2022). A GIS-based landslide susceptibility mapping and variable importance analysis using artificial intelligent training-based methods. *Remote Sensing*, *14*(1), 211.

Zhao, Z., Liu, Z. Y., & Xu, C. (2021). Slope unit-based landslide susceptibility mapping using certainty factor, support vector machine, random forest, CF-SVM and CF-RF models. *Frontiers in Earth science*, *9*, 589630.